\documentclass[manuscript,screen,nonacm]{acmart}%

\pdfoutput=1

\AtBeginDocument{%
  \providecommand\BibTeX{{%
    \normalfont B\kern-0.5em{\scshape i\kern-0.25em b}\kern-0.8em\TeX}}}

\setcopyright{none}%

\usepackage{graphicx}%
\usepackage{caption}
\usepackage{subcaption}

\usepackage{pdflscape}
\usepackage{afterpage}

\usepackage{multirow}%
\usepackage{multicol}
\usepackage{booktabs}%

\usepackage{nicefrac}
\usepackage{tabularx}

\usepackage[inline]{enumitem}

\usepackage{algorithm}
\usepackage{algorithmic}

\usepackage{comment}

\newcommand{\sourcecode}{%
\url{https://github.com/xuanxuanxuan-git/facelift}
}

\graphicspath{{./pdfs/}}

\usepackage{fontawesome}

\usepackage[framemethod=tikz]{mdframed}
\definecolor{mycolor}{rgb}{0.122, 0.435, 0.698}
\definecolor{myblue}{rgb}{0.122, 0.435, 0.698}
\definecolor{mygreen}{rgb}{0.125, 0.525, 0.220}
\definecolor{myyellow}{rgb}{0.588, 0.439, 0.000}
\definecolor{myviolet}{rgb}{0.71764706, 0.40784314, 0.63529412}
\definecolor{myred}{rgb}{0.647, 0.114, 0.165}
\newmdenv[innerlinewidth=0.5pt,roundcorner=4pt,innerleftmargin=6pt,
          innerrightmargin=6pt,innertopmargin=6pt,innerbottommargin=6pt,
          linecolor=mycolor,backgroundcolor=mycolor!25!white]{mybox}
\newmdenv[innerlinewidth=0.5pt,roundcorner=4pt,innerleftmargin=6pt,
          innerrightmargin=6pt,innertopmargin=6pt,innerbottommargin=6pt,
          linecolor=myblue,backgroundcolor=myblue!25!white]{mybluebox}
\newmdenv[innerlinewidth=0.5pt,roundcorner=4pt,innerleftmargin=6pt,
          innerrightmargin=6pt,innertopmargin=6pt,innerbottommargin=6pt,
          linecolor=mygreen,backgroundcolor=mygreen!25!white]{mygreenbox}
\newmdenv[innerlinewidth=0.5pt,roundcorner=4pt,innerleftmargin=6pt,
          innerrightmargin=6pt,innertopmargin=6pt,innerbottommargin=6pt,
          linecolor=myyellow,backgroundcolor=myyellow!25!white]{myyellowbox}
\newmdenv[innerlinewidth=0.5pt,roundcorner=4pt,innerleftmargin=6pt,
          innerrightmargin=6pt,innertopmargin=6pt,innerbottommargin=6pt,
          linecolor=myred,backgroundcolor=myred!25!white]{myredbox}
\newmdenv[innerlinewidth=0.5pt,roundcorner=4pt,innerleftmargin=6pt,
          innerrightmargin=6pt,innertopmargin=6pt,innerbottommargin=6pt,
          linecolor=myviolet,backgroundcolor=myviolet!25!white]{myvioletbox}

\begin{document}

\title[%
Navigating Explanatory Multiverse Through Counterfactual Path Geometry%
]{%
Navigating Explanatory Multiverse Through Counterfactual Path Geometry%
}

\author{Kacper Sokol}
\authornote{Corresponding author.}
\authornote{All authors contributed equally to this research.}
\email{kacper.sokol@inf.ethz.ch}
\orcid{0000-0002-9869-5896}
\affiliation{%
  \institution{Department of Computer Science, ETH Zurich}
  \city{Zurich}
  \country{Switzerland}
}
\affiliation{%
  \institution{ARC Centre of Excellence for Automated Decision-Making and Society, School of Computing Technologies, RMIT University}
  \city{Melbourne}
  \country{Australia}
}

\author{Edward Small}
\authornotemark[2]
\email{edward.small@student.rmit.edu.au}
\orcid{0000-0002-3368-1397}
\affiliation{%
  \institution{ARC Centre of Excellence for Automated Decision-Making and Society, School of Computing Technologies, RMIT University}
  \city{Melbourne}
  \country{Australia}
}

\author{Yueqing Xuan}
\authornotemark[2]
\email{yueqing.xuan@student.rmit.edu.au}
\orcid{0000-0002-9365-8949}
\affiliation{%
  \institution{ARC Centre of Excellence for Automated Decision-Making and Society, School of Computing Technologies, RMIT University}
  \city{Melbourne}
  \country{Australia}
}

\begin{abstract}
Counterfactual explanations are the de facto standard when tasked with interpreting decisions of (opaque) predictive models. %
Their generation is often subject to technical and domain-specific constraints %
that aim to maximise their real-life utility. %
In addition to considering desiderata pertaining to the counterfactual instance itself, guaranteeing existence of a viable path connecting it with the factual data point %
has recently gained relevance. %
While current explainability approaches ensure that the %
steps of such a journey as well as its destination %
adhere to selected constraints, %
they neglect the \emph{multiplicity} of these counterfactual paths. %
To address this shortcoming we introduce the novel concept of \emph{explanatory multiverse} that encompasses all the possible counterfactual journeys. %
We define it using \emph{vector spaces}, %
showing how to navigate, reason about and compare the geometry of %
counterfactual trajectories found within it. %
To this end, we overview their spatial properties %
-- such as affinity, branching, divergence and possible future convergence -- %
and propose an all-in-one metric, called \emph{opportunity potential}, to quantify them. %
Notably, the %
explanatory process %
offered by our method %
grants explainees more agency by allowing them to select counterfactuals %
not only %
based on %
their absolute differences %
but also according to %
the properties of %
their connecting paths. %
To demonstrate real-life flexibility, benefit and efficacy of explanatory multiverse %
we propose its \emph{graph}-based implementation, which we use for %
qualitative and %
quantitative evaluation on %
six tabular and image data sets. %
\end{abstract}


\maketitle

\begin{mybluebox}
\noindent\faGithub\hspace{.2cm}\textbf{Source Code}\quad%
\sourcecode%
\end{mybluebox}
\vspace{.33em}%
\begin{myvioletbox}
\noindent\faFileTextO\hspace{.2cm}\textbf{Published in}\quad%
ECML-PKDD 2025: Journal Track %
(Springer Machine Learning -- \href{https://doi.org/10.1007/s10994-025-06769-2}{10.1007/s10994-025-06769-2}); %
a preliminary version of this work was presented at the \emph{2023 ICML Workshop on Counterfactuals in Minds and Machines} (\href{https://arxiv.org/abs/2306.02786v2}{arXiv:2306.02786v2})
\end{myvioletbox}

\section{Multiplicity of Counterfactual Paths\label{sec:intro}}%

Counterfactuals are the go-to explanations when faced with unintelligible machine learning (ML) classifiers. %
Their appeal -- both to lay and technical audiences -- is grounded in decades of research in social sciences~\cite{miller2019explanation} as well as compliance with various laws and regulations~\cite{wachter2017counterfactual}. %
Fundamentally, counterfactual explanation generation relies on finding an instance whose predicted class is different from that of the factual data point while minimising the distance between the two instances -- a flexible retrieval process that can be tweaked to satisfy bespoke desiderata. %
This versatility makes them more attractive than many other explanation types found in ML such as exemplars, model visualisations, feature attribution or feature importance. %
Additionally, their focus on individual data points, for which they provide precise modification instructions, is their major selling point when needing to explain predictive models deployed in user-facing applications. %

The increasing popularity of counterfactuals and their highly customisable generation mechanism have spurred researchers to incorporate sophisticated technical and social requirements into their retrieval algorithms to better reflect their real-life situatedness. %
Relevant technical aspects include tweaking the fewest possible attributes to guarantee explanation sparsity, or ensuring that counterfactual instances come from the data manifold either by relying on hitherto observed instances~\cite{keane2020good,poyiadzi2020face,van2021interpretable} or by constructing them in dense data regions~\cite{forster2021capturing}. %
Desired social properties reflect real-life constraints imposed by the underlying operational context and data domain, e.g., (im)mutability of certain features, such as date of birth, and directionality of change of others, such as age. %
The mechanisms employed to retrieve these explanations have evolved accordingly: %
from techniques that simply output counterfactual instances~\cite{wachter2017counterfactual,russell2019efficient,romashov2022baycon}, possibly accounting for selected social and technical desiderata as part of optimisation, to methods that %
construct viable paths between factual and counterfactual points%
, sometimes referred to as \emph{algorithmic recourse}~\cite{ustun2019actionable, clark2023trace}, %
ensuring feasibility of this journey and actionability of the prescribed interventions~\cite{downs2020cruds,poyiadzi2020face,forster2021capturing,karimi2021algorithmic}. %

All of these properties aid in generating admissible counterfactuals, but they do not offer guidance on how to discriminate between them. %
Available explainers tend to output multiple counterfactuals whose differences, as it stands, can only be captured through: high-level desiderata, e.g., chronology and coherence; quantitative metrics, e.g., completeness, proximity and neighbourhood density; or qualitative assessment, e.g., user trust and confidence~\cite{sokol2020explainability,keane2021if,small2023helpful,sokol2024what,xuan2023can}. %
In principle, explanation plurality is desirable since it facilitates (interactive) personalisation by offering versatile sets of actions to be taken by the explainees instead of just the most optimal collection of steps (according to a predefined objective), thus catering to %
unique needs and expectations of diverse audiences~\cite{sokol2018glass,sokol2020one}. %
Nonetheless, this multiplicity is also problematic as without incorporating domain-specific knowledge, collecting user input or relying on generic ``goodness'' heuristics and criteria to filter out redundant explanations -- which aspects remain open challenges in themselves -- they are likely to overwhelm the explainees~\cite{keane2021if,herm2023impact}. %

While comparing counterfactual instances based on their diversity as well as overall feasibility and actionability, among other properties, may inform their pruning, considering the (geometric) relation between the paths leading to them could prove more potent. %
Current literature treats counterfactual explanations as \emph{independent} and disregards both their lineage as well as the process that transforms the factual data point into the counterfactual instance, which may unravel over an extended time period~\cite{barocas2020hidden,verma2020counterfactual,beretta2023importance,de2024time}. %
The \emph{spatial relation} between counterfactual paths is also overlooked, whether captured by their direction, length or number of discrete steps (including alignment and size thereof). %
Additionally, while the feasibility of attribute tweaks, i.e., feature actionability or mutability, is often considered, the direction and magnitude of these changes are largely neglected, e.g., age can only increase at a fixed rate. %
Modelling such aspects of counterfactual paths is bound to offer domain-independent, spatially-informed heuristics that reduce the number of explanations for users to consider, grant them agency, inform their decision-making and support reasoning as well as forward planning. %

\begin{figure}[t]
    \centering
    \includegraphics[trim={10pt 10pt 10pt 10pt},clip,width=.450\linewidth]{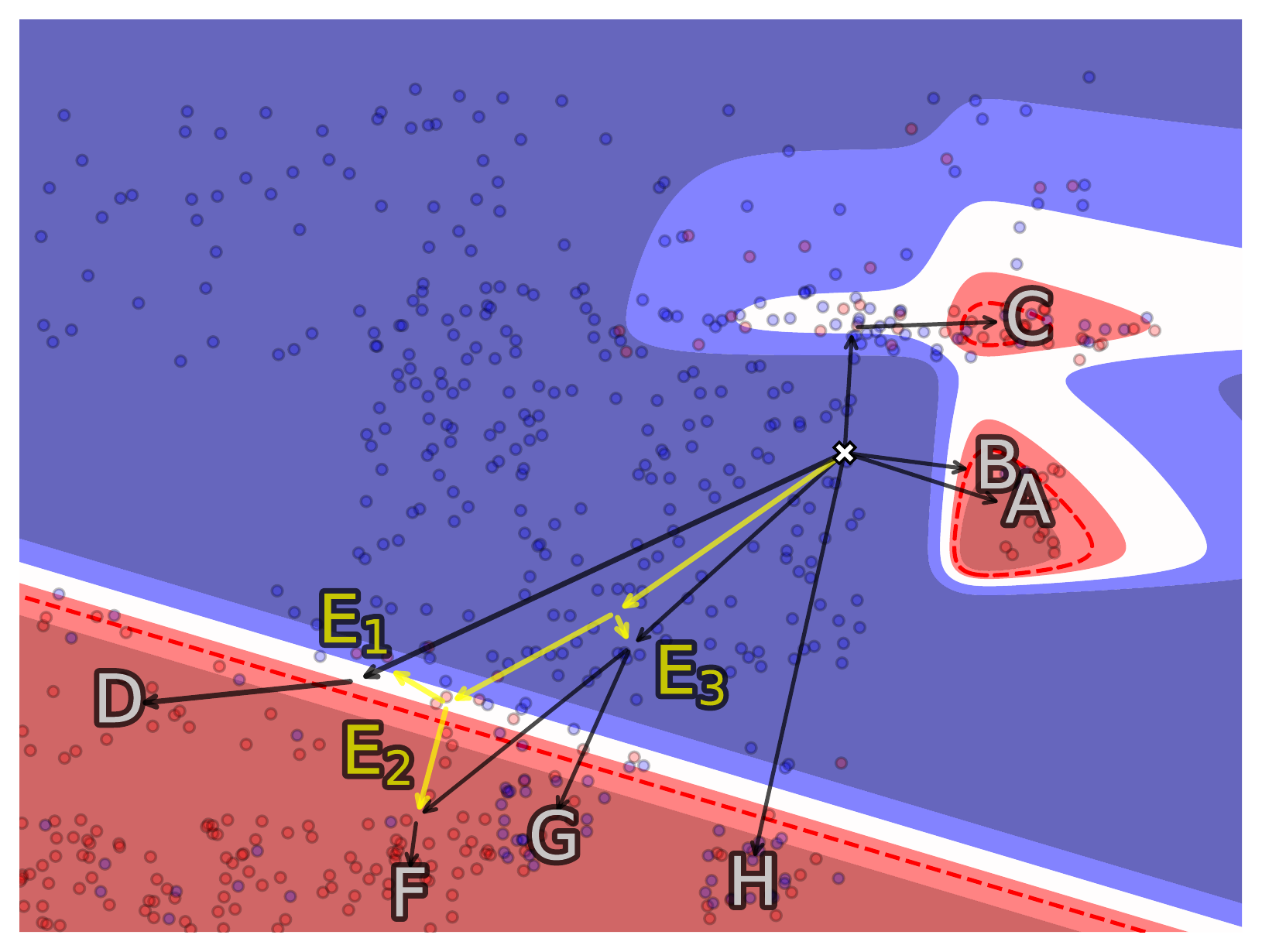}%
    \caption{Example of \emph{explanatory multiverse} constructed for tabular data with two continuous (numerical) features. It demonstrates various types of \emph{counterfactual path geometry} such as the \emph{affinity}, \emph{branching}, \emph{divergence} and \emph{convergence} of these explanations. %
    Each journey terminates in a (possibly the same or similar) counterfactual explanation but the characteristics of the steps leading there make some explanations more attractive targets, e.g., by giving the explainee more agency through multiple actionable choices towards the end of the path.}%
    \label{fig:toy_multiverse}%
\end{figure}

The role of geometry in (counterfactual) explainability is captured by Figure~\ref{fig:toy_multiverse}; it demonstrates the diverse characteristics of counterfactual paths for a two-dimensional toy data set with continuous numerical features. %
When considered in \emph{isolation}, these paths have the following properties: %
\begin{description}[labelindent=0cm,labelwidth=1em,labelsep=1em,align=left,font=\bf]%
\setlength\itemsep{0em}
    \item [A] is short and leads to a high-confidence region, but it lacks data along its journey, which signals infeasibility; %
    \item [B] while shorter, it terminates close to a decision boundary, thus carries high uncertainty; %
    \item [C] addresses the shortcomings of A, but it lands in an area of high instability (compared to D, E\textsubscript{i}, F, G \& H); %
    \item [G \& H] also do not exhibit the deficiencies of A, but they are located in a region with a high error rate; %
    \item [D \& F] have all the desired properties, but they are the most distant; and %
    \item [E\textsubscript{i}] are feasible, but they are \emph{incomplete} by themselves. %
\end{description}
While such considerations have become commonplace in the literature, they treat each destination and the path leading to it as independent, thus forgoing the benefit of accounting for their spatial relation such as affinity, branching, divergence or convergence. %
For example, accessing D, F \& G via E\textsubscript{i} elongates the journey leading to these counterfactuals, making them less attractive based on current desiderata; %
nonetheless, such a path maximises the agency of explainees by providing them with multiple recourse choices along a \emph{unified} trajectory. %

Reasoning about the properties of and comparing counterfactual paths are intuitive in two-dimensional spaces -- as demonstrated by Figure~\ref{fig:toy_multiverse} -- but doing so in higher dimensions requires a more principled approach. %
To this end, we propose the notion of \emph{explanatory multiverse}, grounding it in the current literature (Section~\ref{sec:prelim}); %
this novel explainability framework embraces the multiplicity of counterfactuals and captures the \emph{geometry}, i.e., spatial dependence, of journeys leading to them. %
Our conceptualisation accounts %
for technical, \emph{spatially-unaware} desiderata prevalent in the literature -- e.g., plausibility, path length, number of steps, sparsity, magnitude of attribute change and feature actionability -- and defines novel, \emph{spatially-aware} properties such as branching delay, branching factor, (change in) path directionality and affinity between journeys. %

Specifically, we formalise explanatory multiverse with \emph{vector spaces}, in which counterfactual paths are composed of connected vectors (Section~\ref{sec:vector}); %
this representation is a natural fit for our framework and allows us to effectively communicate the core (technical) concepts underlying our ideas -- refer to Figure~\ref{fig:toy_multiverse}. %
Vector spaces can be built upon data density estimates (in a model-agnostic fashion) or rely directly on gradients if the underlying model can provide this information. %
While our formalisation presupposes continuous feature spaces %
-- for which it is highly suitable -- %
it can also handle discrete attributes given appropriate pre-processing, e.g., one-hoc encoding. %
In addition to imbuing counterfactuals with spatial awareness, explanatory multiverse reduces the number of admissible explanations by \emph{collapsing} paths based on their affinity and \emph{pruning} them through our spatially-aware desiderata. %
To quantify these properties
we formalise an all-in-one metric called \emph{opportunity potential}. %

To demonstrate the practical aspects and real-life advantages of explanatory multiverse we introduce its (somewhat limited) %
implementation that relies on (directed) \emph{graphs} (Section~\ref{sec:graph}). %
These are built from data based on a predefined distance metric and are highly suitable for predominantly discrete feature spaces; %
here, counterfactuals are retrieved through pathfinding. %
Specifically, we experiment on three tabular -- German Credit~\cite{statlog}, Adult Income~\citep{becker1996adult} and Credit Default~\citep{credit_default} -- %
and three image -- MNIST~\cite{lecun1998mnist} as well as BreastMNIST and PneumoniaMNIST (from the MedMNIST collection~\citep{medmnistv1}) -- data set (Section~\ref{sec:experiment}). %
Our results %
demonstrate the flexibility, benefit and efficacy of our approach through examples and numerical evaluation. %
We invite others to embark on this novel counterfactual explainability research journey by releasing a Python package, called \textsc{FACElift}, that implements our methods.\footnote{\sourcecode} %

From the explainees' perspective, our approach helps them to navigate explanatory multiverse and grants them initiative and agency, empowering meaningful exploration, customisation and personalisation of counterfactuals. %
For example, by choosing a path recommended based on a high number of diverse explanations accessible along it -- refer to segments E\textsubscript{i} in Figure~\ref{fig:toy_multiverse}, which have high opportunity potential -- the user is given ample chance to receive the desired outcome while applying a consistent set of changes, thus reducing the number of u-turns and backtracking arising in the process. %
Explanatory multiverse is also compatible with human-in-the-loop, interactive explainability~\cite{sokol2018glass,sokol2020one,keenan2023mind} and %
a relatively recent decision-support model of explainability, which relies on co-construction of explanations and complements the more ubiquitous ``prediction justification'' paradigm~\cite{miller2023explainable}. %
We explore and discuss these concepts further in Section~\ref{sec:discussion}, which also outlines future work, before concluding the paper in Section~\ref{sec:conclusion}. %

\section{Preliminaries\label{sec:prelim}}

Before we formalise explanatory multiverse desiderata (Section~\ref{sec:prelim:desiderata}), we position the underlying idea in the current literature (Section~\ref{sec:prelim:related}) and summarise the notation (Section~\ref{sec:prelim:notation}) used to introduce %
its vector-based conceptualisation and graph-based implementation. %
Throughout this paper we assume that we are given a (state-of-the-art) explainer that generates counterfactuals along with steps leading to them, such as FACE~\cite{poyiadzi2020face} or any other similar technique~\cite{ustun2019actionable,karimi2021algorithmic}. %
Our ideas are compatible with explainers that follow actual data instances as well as those that operate directly on feature spaces (e.g., relying on data distribution density). %

\subsection{Related Work\label{sec:prelim:related}}

Existing counterfactual explainers account for multiple desiderata pertaining to the counterfactual instances themselves -- i.e., \emph{spatially-unaware} properties -- to ensure their real-life practicality. %
The most common requirement concerns the \emph{distance} between the factual and counterfactual instances as well as the number of features being tweaked, striving for the smallest possible distance and number of affected attributes~\cite{ tolomei2017interpretable,wachter2017counterfactual}. %
Another desideratum accounts for (domain-specific) constraints with respect to feature alterations, specifically \emph{(im)mutability} of attributes, direction and rate of their change as well as their \emph{actionability} from a user's perspective; e.g., some tweaks are irreversible, some can be implemented by explainees, and others -- mutable but non-actionable -- are properties of the environment~\cite{ustun2019actionable,karimi2021algorithmic}. %
\emph{Plausibility} of the counterfactual data point is also considered. %
It requires the explanations to be: %
\begin{enumerate*}[label=(\arabic*)]
\item
\emph{feasible} according to the underlying data distribution, thus come from the data manifold, enforced either through density constraints or by following pre-existing instances; and %
\item
of \emph{high confidence}, i.e., robust, in relation to the explained predictive model~\cite{downs2020cruds,pawelczyk2020learning,poyiadzi2020face,van2021interpretable}. %
\end{enumerate*}
The \emph{multiplicity} of admissible counterfactual explanations is their least explored property; %
explainers should aim to output a comprehensive subset of instances that are the \emph{most diverse}, \emph{least similar} as well as \emph{highly representative}~\cite{mothilal2020explaining,keane2021if,laugel2023achieving}. %

While important, these desiderata are only concerned with the counterfactual instances themselves, thus overlooking the properties of paths connecting them with the data points being explained. %
Ignoring the journey between the factual and counterfactual instances is synonymous with assuming that a user can implement such changes simultaneously and instantaneously, whereas in reality this process often involves multiple discrete actions spread over time~\cite{barocas2020hidden}. %
Explainers such as FACE can generate a counterfactual path as a sequence of steps based on training data points, but methods capable of outputting explanations in this format are scarce~\cite{ramakrishnan2020synthesizing,kanamori2021ordered,verma2022amortized}. %
These approaches, however, treat path-based counterfactual explanations as independent -- ignoring the useful information carried by their spatial relation -- which gap creates a need for optimisation objectives and evaluation metrics that explicitly capture various characteristics of counterfactual paths. %
As a first step, counterfactual instance-based metrics can be adapted to this setting, e.g., a path length and its number of steps as well as actionability and plausibility of the intermediate instances that constitute it. %
Nonetheless, we still lack desiderata that capture the continuity of steps given by counterfactual paths, multiplicity thereof, and their geometrical relation. %
Such properties become especially important for real-life implementation of algorithmic recourse that may stretch over time, in which case a path may become infeasible part way through its enactment, e.g., due to uncontrollable factors, preventing its completion and requiring a transition to an alternative route. %

\subsection{Notation\label{sec:prelim:notation}}

We denote the $m$-dimensional input space as $\mathcal{X}$, with the explained (factual) instance given by $\mathring{x}$ and the explanation (counterfactual) data point by $\check{x}$. %
These instances are considered in relation to a predictive model $f : \mathcal{X} \mapsto \mathcal{Y}$, where $\mathcal{Y}$ is the space of possible classes; %
therefore, $f(\mathring{x}) \neq f(\check{x})$ and the prediction of the former data point $f(\mathring{x}) = \mathring{y} \in \mathcal{Y}$ is less desirable than that of the latter $f(\check{x}) = \check{y} \in \mathcal{Y}$. %
$\widetilde{f}$ refers to the probabilistic realisation of $f$ that outputs the probability of the desired class. %
Vector $p$-norms, which provide a measurement of length, are defined as %
$
        \lVert v \rVert_p = \left(\sum_{i=1}^n \lvert v_i\rvert^p\right)^{\frac{1}{p}} %
$.
Distance functions, e.g., on the input space, are denoted by $d : \mathcal{X} \times \mathcal{X} \mapsto \mathbb{R}^+$. %

To discover $\check{x}$ and the \emph{steps} necessary to transform $\mathring{x}$ into this instance, we apply a state-of-the-art, path-based explainer -- see Section~\ref{sec:prelim:related} for the properties expected of it -- that generates (multiple) counterfactuals along with their journeys. %
Each such explanation $Z^{[k]}$, where $k$ is its index, is represented by an $m \times n$ matrix whose $n$ columns $z^{[k]}_i \in \mathcal{X}$ capture the sequence of steps in the $m$-dimensional input space, i.e., %
$ Z^{[k]} = [ z_1^{[k]} \; \cdots \; z_n^{[k]} ] $,
such that if we start at the factual point $\mathring{x}$, we end at the counterfactual instance $\check{x}$ like so:
$
        \check{x} = \mathring{x} + \sum_{i=1}^{n} z_i^{[k]}
$.
In the \emph{vector-based} conceptualisation of explanatory multiverse,
each step can %
\emph{optionally} be %
discounted by %
a weight factor $w_i$; %
this parameterisation allows imposing %
user-specified properties onto the counterfactual path. %
For example, if the explainees want to prioritise early choices as to ensure that a large number of target counterfactuals remains available for the first couple of steps, %
their preference can be captured by the weight vector $w = [w_1 \; \cdots \; w_n]$ adhering to the following constraints: %
$\lVert w \rVert_2 = 1$ and $w_1 \geq w_2 \geq \cdots \geq w_{n-1} \geq w_n$. %

For the \emph{graph-based} implementation of explanatory multiverse, %
we take $G = (V, E)$ to be a (directed) graph with vertices $V$ and arcs (directed edges) $E$. %
Each vertex $v_i \in V$ corresponds to a data point $x_i \in \mathcal{X}$ in the input space, i.e., $v_i \equiv x_i$; %
these are connected with (directed) edges $e_{i,j} \in E$ that leave $v_i$ and enter $v_{j}$. %
Assuming that $v_1$ represents the explained instance $\mathring{x}$, thus the starting point of a path, and $v_n$ is the target counterfactual data point $\check{x}$, thus where the journey terminates, we can identify multiple paths $Z^{[k]}$ connecting them. %
Here, the columns $z_i^{[k]}$ of the $m \times n$ matrix $Z^{[k]}$ representing an $n$-step explanatory journey store a sequence of vertices $v_i$ along this path, i.e., $z_i^{[k]} \equiv v_i$, with $z_1^{[k]} \equiv \mathring{x}$ and $z_n^{[k]} \equiv \check{x}$. %
Such a journey can alternatively be denoted by the corresponding sequence of edges $E_{1, n} = [e_{1,2} \; \cdots \; e_{n-1,n}]$ -- where $e_{i,i+1} \in E$ -- %
that encode %
the properties of each transition (akin to the weights $w_i$ used in the vector space conceptualisation). %

\subsection{Spatially-aware Desiderata\label{sec:prelim:desiderata}}%

Our \emph{explanatory multiverse} framework %
builds on top of the \emph{spatially-unaware} properties discussed in Section~\ref{sec:prelim:related} -- which it inherits by relying on state-of-the-art, path-based counterfactual explainers -- and extends these desiderata with three novel, \emph{spatially-aware} properties. %
\begin{description}[labelindent=0cm,labelwidth=1em,labelsep=1em,align=left,font=\bf]%
\setlength\itemsep{0em}
    \item [Agency] %
    captures the number of choices -- leading to \emph{diverse} counterfactuals -- available to explainees as they traverse explanatory paths. %
    High agency offers a selection of explanations and stimulates \emph{user initiative}. %
    It can be measured as \emph{branching factor}, i.e., the number of paths leading to representative explanations accessible at any given step. %
    \item [Loss of Opportunity] %
    encompasses the \emph{incompatibility} of subsets of explanations that emerges as a consequence of implementing changes prescribed by steps along a counterfactual path. %
    For example, moving towards one explanation may require backtracking the steps taken thus far -- which may be impossible due to strict directionality of change imposed on relevant features -- to arrive at a different, equally suitable counterfactual. %
    Such a \emph{loss of opportunity} can be measured by (a decrease in) the proportion of counterfactuals reachable after taking a step without the need of backtracking, which can be quantified by the degree of change in \emph{path directionality} (vectors) or \emph{vertex inaccessibility} (graphs). %
    \item [Choice Complexity] %
    encapsulates the influence of explainees' decisions to follow a specific counterfactual path on the availability of diverse alternative explanations. %
    It can be understood as \emph{loss of opportunity} accumulated along different explanatory trajectories. %
    For example, among otherwise equivalent paths, following those whose \emph{branching is delayed} reduces early commitment to a particular set of explanations. %
    It can be measured by the distance (or number of steps and their magnitude) between the factual data point and the earliest consequential point of \emph{counterfactual path divergence}. %
\end{description}

To illustrate the benefits of navigating counterfactuals through the lens of explanatory multiverse, %
consider an individual who is currently ineligible for a loan but who can follow different sets of actions to eventually receive it. %
Choosing one of these paths may preclude others, e.g., getting a full time job is incompatible with pursuing higher education -- a scenario captured by \emph{loss of opportunity}. %
Implementing the individual actions that are the most universal, thus shared across many paths to a successful loan application, allows to delay the funnelling towards specific sequences of actions -- a benefit of accounting for \emph{choice complexity}. %
In general, the landscape of available paths can be more easily navigated by taking actions that do not limit explainees' choices -- a prime example of \emph{agency}. %
To capture these desiderata we propose a novel all-in-one metric, called \emph{opportunity potential}, %
and formalise it for the vector conceptualisation and graph implementation of explanatory multiverse respectively in Sections~\ref{sec:vector} and \ref{sec:graph}. %
It quantifies the geometrical alignment of one path with respect to another, implicitly encapsulating all of the above properties as demonstrated by the experiments reported in Section~\ref{sec:experiment}. %
Notably, we can strive for all of these desiderata simultaneously, weighting some in favour of others if necessary, which we explore further in Section~\ref{sec:discussion}. %

\section{Conceptualising Explanatory Multiverse with Vector Spaces\label{sec:vector}}%

It should be clear by now that offering each admissible counterfactual as an independent sequence of steps and ordering these explanations purely based on their overall length fails to fully capture some fundamental, human-centred properties, e.g., agency. %
We must therefore seek strategies to compare the geometry of each counterfactual path, but such a task comes with challenges of its own. %
In this section we formalise explanatory multiverse through vector spaces and %
introduce a comprehensive (theoretical) framework that covers its core concepts. %

\paragraph{Comparing Journeys of Varying Length}
Counterfactual paths may differ in their overall length, number of steps, individual magnitude thereof and the like. %
For example, juxtapose paths B and F in Figure~\ref{fig:toy_multiverse}. %
The former is composed of fewer steps and is much shorter than the latter, making the direct geometrical comparison via their original components infeasible -- we could only do so up to the number of steps of the shorter of the two paths (assuming that these steps themselves are of similar length). %
We address this challenge by %
comparing counterfactual journeys through their normalised relative sections. %

Given two paths $Z^{[a]} \in \mathbb{R}^{m\times n_1}$ and $Z^{[b]} \in \mathbb{R}^{m\times n_2}$ differing in their number of steps, i.e., $n_1 \neq n_2$, we encode them with an equal number of vectors $o$ such that $\bar{Z}^{[a]}, \bar{Z}^{[b]} \in \mathbb{R}^{m\times o}$ are normalised journeys. %
To this end, we define a function %
\begin{equation*}
    c_L (Z) = \sum_{i=1}^{n} \lVert z_i \rVert_2
\end{equation*}
that provides us with the total length of a counterfactual path $Z \in \mathbb{R}^{m \times n}$. %
We then select the number of steps $o$ into which we partition each journey -- these serve as comparison points for paths of differing length. %
Therefore, the $j$\textsuperscript{th} step $\bar{z}_j$ of a \emph{normalised} counterfactual journey $\bar{Z}\in\mathbb{R}^{m\times o}$ is %
$$
    \bar{z}_j = \sum_{i=1}^{o} 
    \delta_{z_i} \left( \frac{j}{o} c_L(Z) -  \sum_{l=1}^i \lVert z_l \rVert_2 \right) z_i
    \text{~,}
$$
where
$$
    \delta_{z_i} (\zeta) =
    \begin{cases}
        0 &\text{if} \; \zeta \leq 0 \\
        \frac{\zeta}{\lVert z_i \rVert_2} &\text{if} \; 0 < \zeta \leq \lVert z_i \rVert_2  \\
        1 &\text{otherwise.} %
    \end{cases}
$$

\begin{figure*}[t]
     \centering
     \begin{subfigure}[b]{0.325\textwidth}%
         \centering
         \includegraphics[trim={10pt 10pt 10pt 10pt},clip,width=\textwidth]{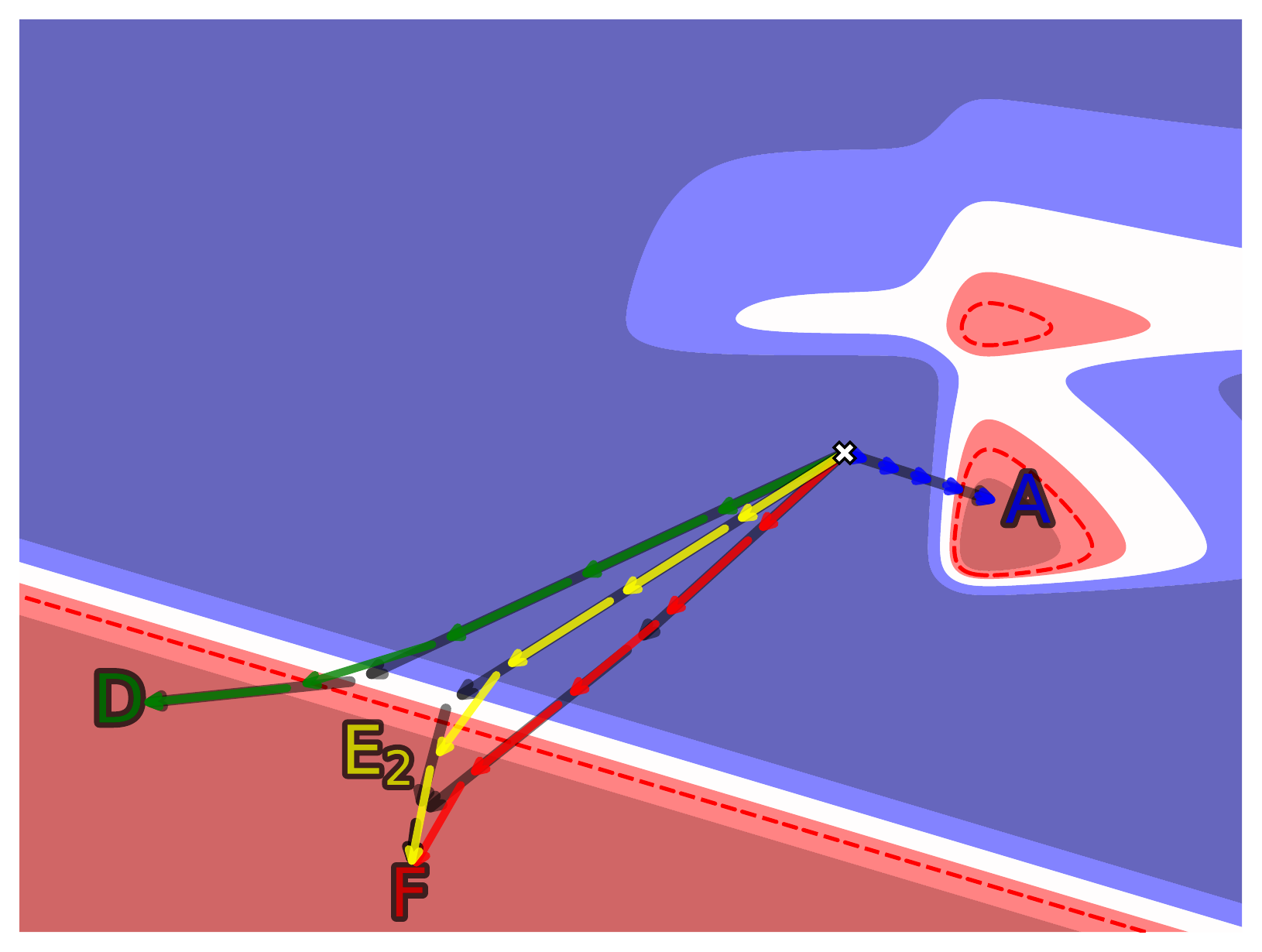} %
         \caption{$o=5$.\label{fig:kis5}}
     \end{subfigure}
     \hfill
     \begin{subfigure}[b]{0.325\textwidth}%
         \centering
         \includegraphics[trim={10pt 10pt 10pt 10pt},clip,width=\textwidth]{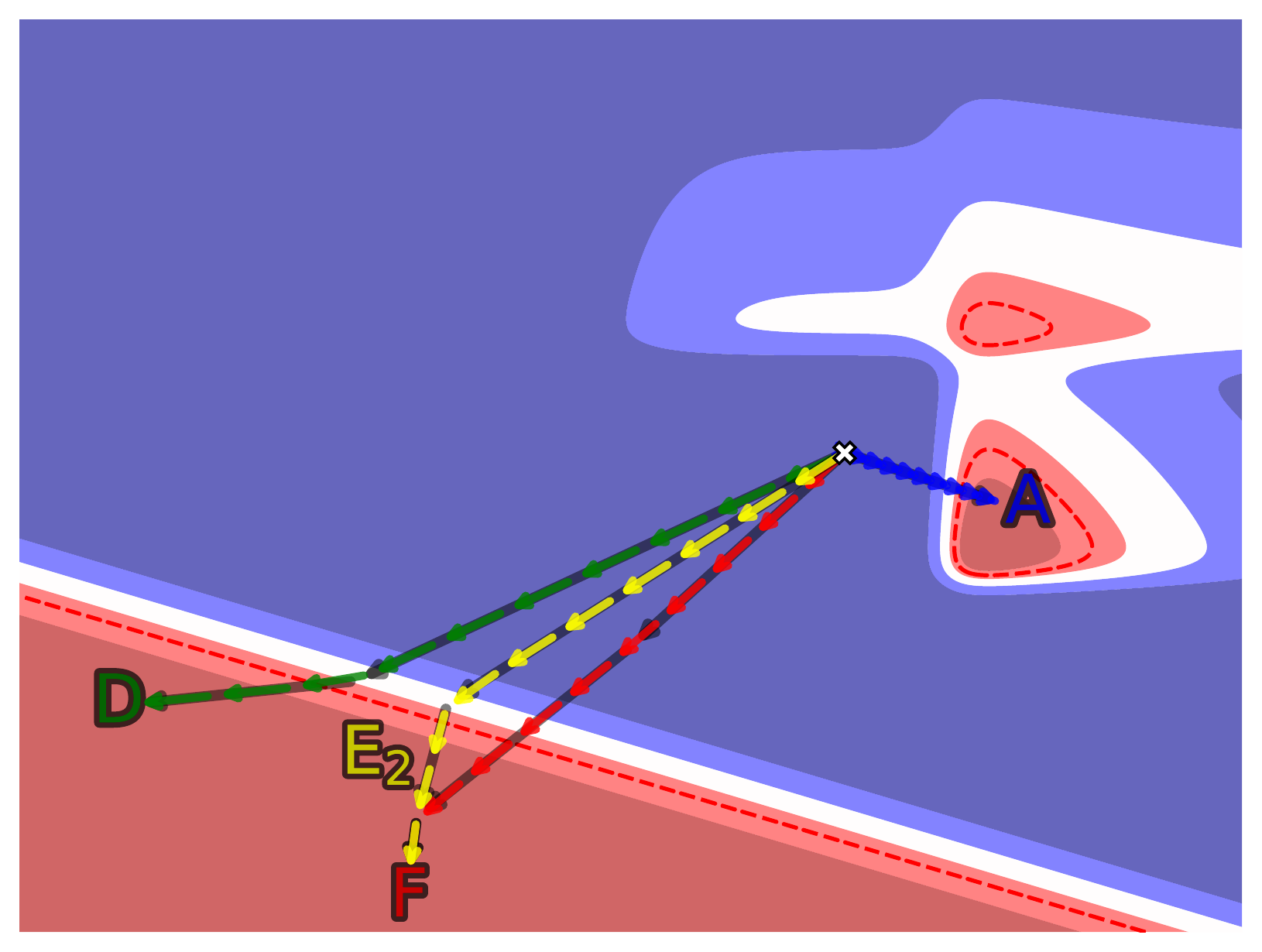} %
         \caption{$o=10$.\label{fig:kis10}}
     \end{subfigure}
     \hfill
     \begin{subfigure}[b]{0.325\textwidth}%
         \centering
         \includegraphics[trim={10pt 10pt 10pt 10pt},clip,width=\textwidth]{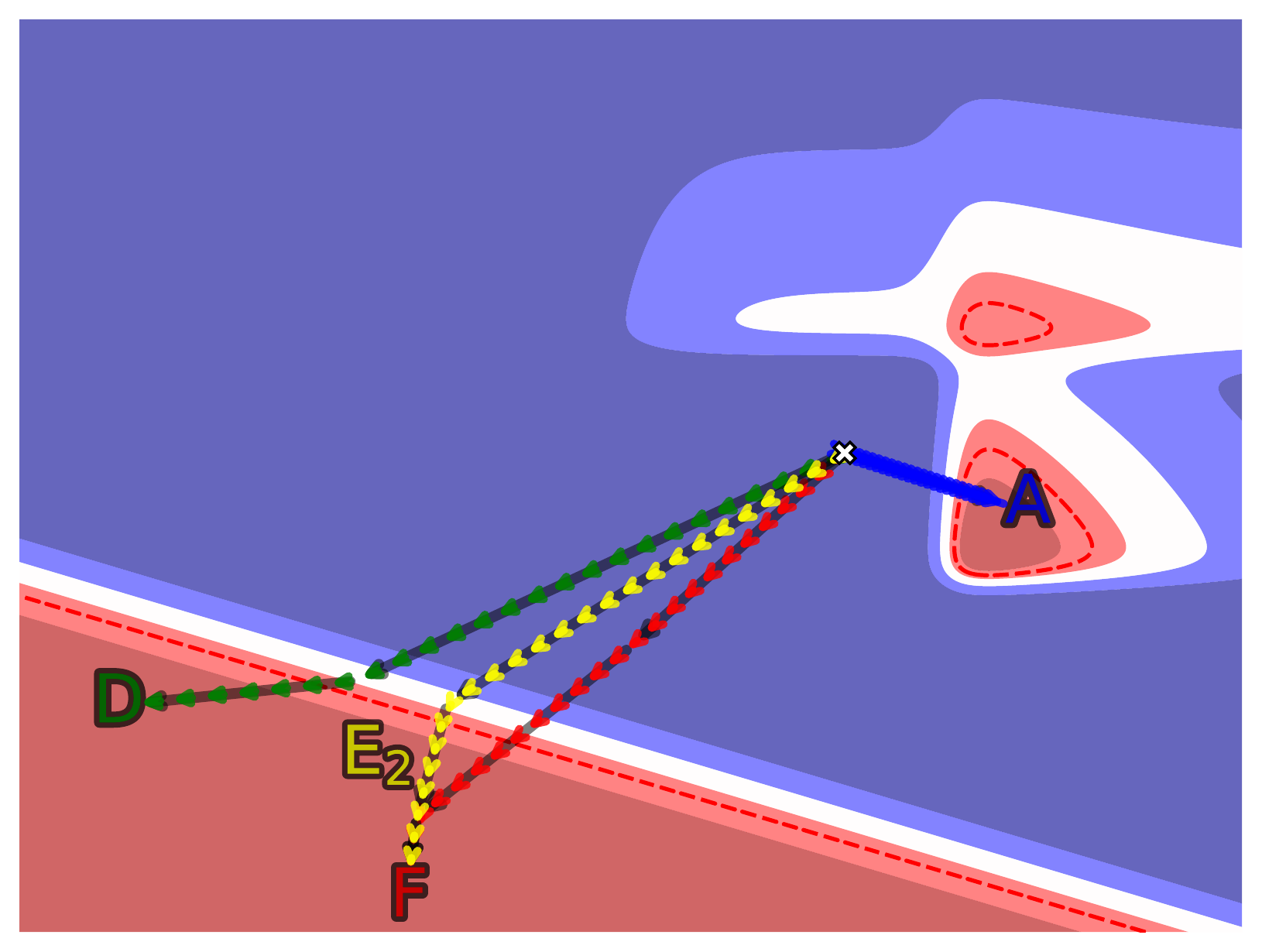} %
         \caption{$o=25$.\label{fig:kis25}}
     \end{subfigure}
          \caption{%
     Demonstration of how the number of vectors $o$ into which a path is split %
     affects counterfactual trajectories. %
     This parameter must be carefully selected, which may require domain knowledge and familiarity with the underlying data. %
     }%
     \label{fig:osplit}%
\end{figure*}

This procedure -- demonstrated in Figure~\ref{fig:osplit} and codified by Algorithm~\ref{alg:cap} -- %
allows us to directly compare $\bar{Z}$ with any other path normalised to the same number of steps~$o$. %
While necessary, such normalisation can %
negatively impact feasibility of a counterfactual journey; %
this can be seen in Figure~\ref{fig:kis5} for the yellow path, which uses a ``shortcut'' when crossing the decision boundary. %
This behaviour can nonetheless be easily monitored with off-the-shelf feasibility metrics and $o$ can then be adjusted accordingly to mitigate any unintended artefacts of the normalisation. %

\begin{algorithm}[t]
\caption{Split counterfactual path $Z$ into $o$ vectors.}%
\label{alg:cap}
\begin{algorithmic}[1]
    \REQUIRE
        counterfactual path $Z \in \mathbb{R}^{m \times n}$; %
        number of partitions $o \in \mathbb{Z}^+$. %
    \ENSURE
        counterfactual path $\bar{Z} \in \mathbb{R}^{m \times o}$ with $o$ steps. %
    \FOR{$j \gets 1$; \quad $j \gets j + 1$; \quad $j \leq o$}
        \STATE $\beta \gets \frac{j}{o} c_L(Z)$ %
        \COMMENT{split path's total length}%
        \STATE $\tau \gets \mathbf{0}$ %
        \FOR{$i \gets 1$; \quad $i \gets i + 1$; \quad $i \leq n$}
             \IF{$\beta < 0$}
                 \STATE \emph{continue} %
             \ELSIF{$0 \leq \beta \leq \lVert z_i \rVert_2$}
                 \STATE $\tau \gets \tau + \frac{\beta}{\lVert z_i \rVert_2}z_i$
             \ELSE
                 \STATE $\tau \gets \tau + z_i$
             \ENDIF
            \STATE $\beta \gets \beta - \lVert z_i \rVert_2$
        \ENDFOR
        \STATE $\bar{z}_j \gets \tau$
    \ENDFOR
\end{algorithmic}
\end{algorithm}

\begin{figure*}[t]
     \centering
         \includegraphics[trim={10pt 10pt 10pt 10pt},clip,height=3.465cm]{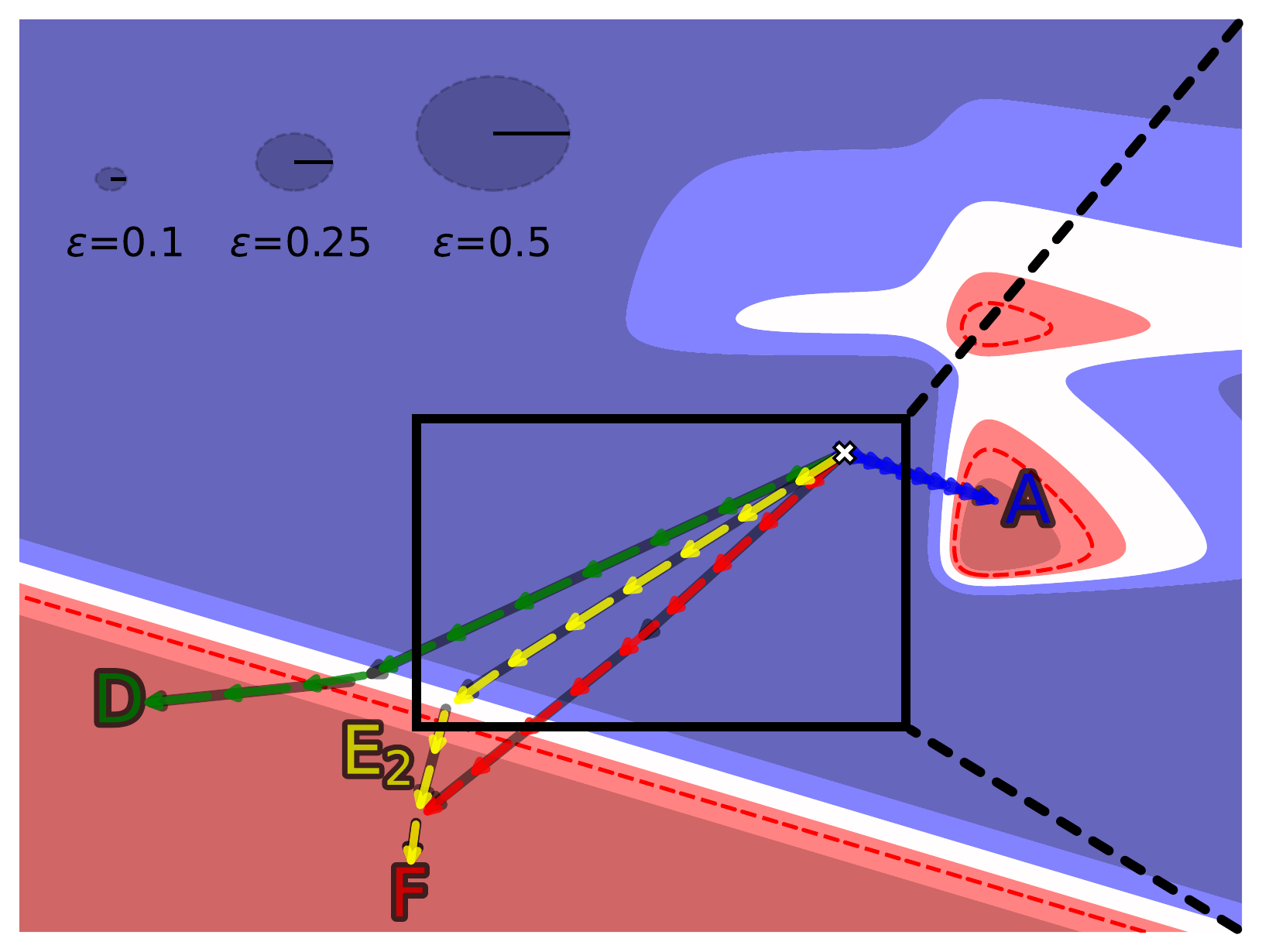} %
     \hspace{-0.26cm}%
         \includegraphics[trim={10pt 10pt 10pt 10pt},clip,height=3.465cm]{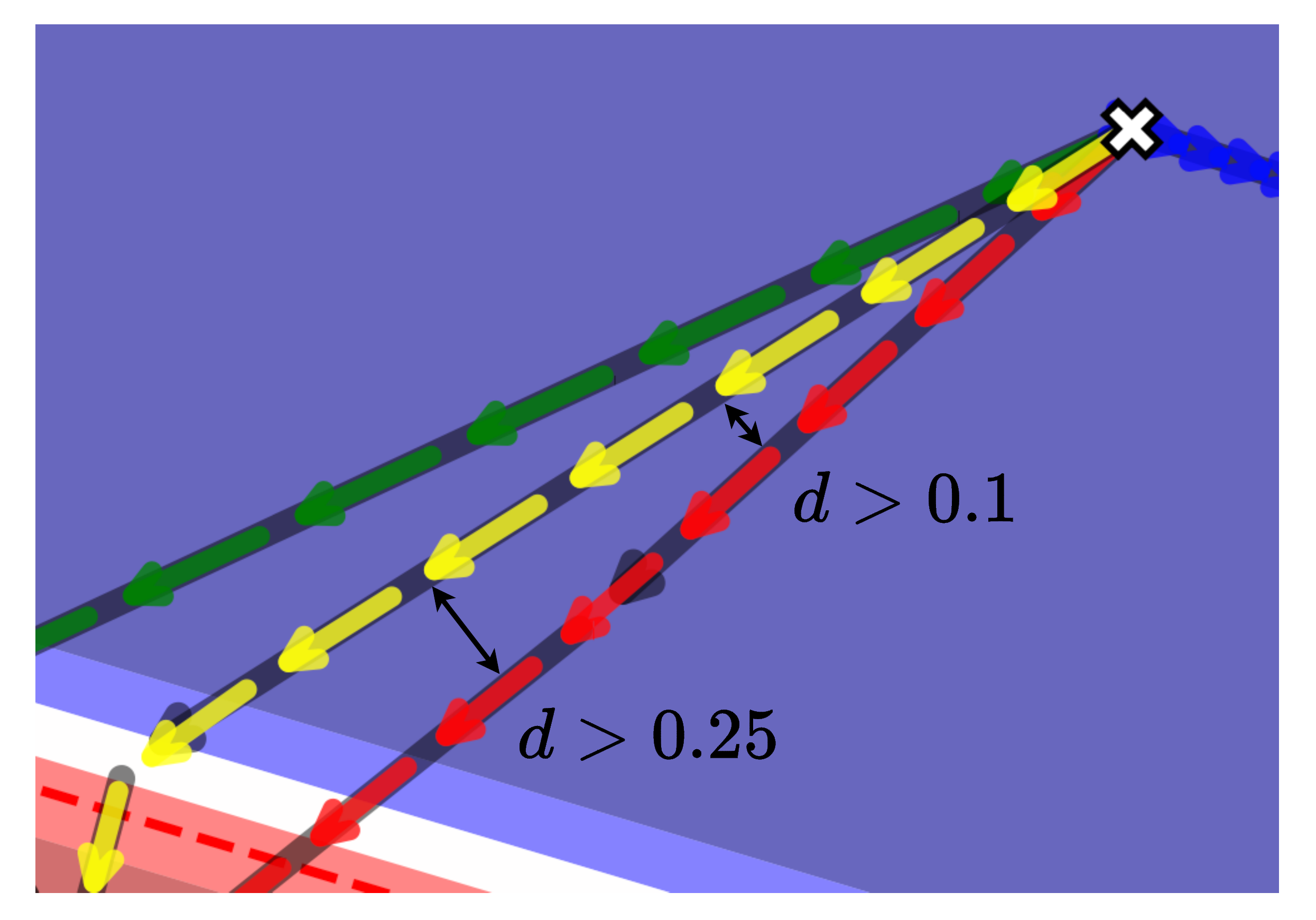} %
     \caption{%
     Demonstration of how the
     branching (i.e., divergence) threshold $\epsilon$
     affects counterfactual trajectories and \emph{branching points}. %
     Here, the paths are split into %
     $o=10$ vectors; %
     $\epsilon$ must be carefully selected, which may require domain knowledge and familiarity with the underlying data. %
     In this example, %
     if $\epsilon=0.1$, the yellow path \emph{diverges} from the red path after its third step %
     since their distance $d$ exceeds $\epsilon$. %
     When $\epsilon=0.25$, the divergence occurs after the yellow path's fifth step. %
     If $\epsilon \geq 0.5$, the two paths do not diverge given that $\epsilon$ is greater than the distance separating them at every point along their way}. %
     \label{fig:epsilonbranch}%
\end{figure*}

\paragraph{Identifying Branching Points}
We can calculate the proximity between two counterfactual paths $\bar{Z}^{[a]} , \bar{Z}^{[b]} \in \mathbb{R}^{m \times o}$ at all points along $\bar{Z}^{[a]}$, therefore identify the precise location of their divergence. %
To this end, we compute the minimum distance between the path $\bar{Z}^{[b]}$ and the $i$\textsuperscript{th} point $\bar{z}_i^{[a]}$ on the path $\bar{Z}^{[a]}$ with %
\begin{equation*}
    \bar{z}_i^{[a \vert b]} =
    d_S( \bar{Z}^{[b]} - z^{[a]}_i \mathbf{1}_o^T ) 
    \text{~,}
\end{equation*}
where
\begin{equation*}
    d_S(Z) =
    \min_{1 \leq j \leq o} \lVert z_j \rVert_2 =
    \min_{1 \leq j \leq o} \sqrt{\sum_{i=1}^m \lvert z_{i,j}\rvert^2}
\end{equation*}
and $\mathbf{1}_o^T = [1 \; \cdots \; 1]$ such that $\vert \mathbf{1}_o \vert = o$. %
Next, we define a \emph{branching threshold} $\epsilon > 0$ such that two paths are considered to have separated at a \emph{divergence point} $o^\star$ where $\bar{z}_i^{[a \vert b]}$ first exceeds the threshold $\epsilon$, i.e., %
$$
o^\star = \min(i) \quad \text{s.t.} \quad \bar{z}_i^{[a \vert b]} > \epsilon
\text{~.}
$$
We can then denote the proportion along the journey $\bar{Z}^{[a]}$ before it branches away from $\bar{Z}^{[b]}$ as $\frac{o^\star - 1}{o}$. %
This procedure is demonstrated in Figure~\ref{fig:epsilonbranch} and captured by Algorithm~\ref{alg:branching}. %

\begin{algorithm}[t]
\caption{Identify branching point $o^\star$.}
\label{alg:branching}
\begin{algorithmic}[1]
    \REQUIRE
        normalised counterfactual paths $\bar{Z}^{[a]}, \bar{Z}^{[b]} \in \mathbb{R}^{m\times o}$; %
        branching threshold $\epsilon > 0$. %
    \ENSURE
        divergence point $o^\star$. %
    \FOR{$i \gets 1$; \quad $i \gets i + 1$; \quad $i \leq o$}
        \IF{$\bar{z}_i^{[a \vert b]} > \epsilon$} %
             \STATE $o^\star \gets i$
             \STATE \emph{break} %
        \ENDIF
    \ENDFOR
\end{algorithmic}
\end{algorithm}

\paragraph{Direction Difference Between Paths}
After normalising two counterfactual paths to have the same number of steps, i.e., $\bar{Z}^{[a]} , \bar{Z}^{[b]} \in \mathbb{R}^{m \times o}$, we can compute direction difference between them. %
Popular distance metrics, e.g., the Euclidean norm, can be adapted to this end: %
$$
\begin{aligned}
    d_E( \bar{Z}^{[a]} , \bar{Z}^{[b]} ) &= \sum_{j=1}^o w_i \rVert \bar{z}^{[a]}_{i,j} - \bar{z}^{[b]}_{i,j} \rVert_2 \\
    &= \sum_{j=1}^o w_i \sqrt{\sum_{i=1}^m ( \bar{z}^{[a]}_{i,j} - \bar{z}^{[b]}_{i,j} )^2}
\text{~,}
\end{aligned}
$$
where $d_E : Z \times Z \mapsto \mathbb{R}^+$ and the weight vector $w$, with $w \in \mathbb{R}^o$, is as outlined in Section~\ref{sec:prelim:notation}. %

$d_E$ therefore offers a measure of directional separation between two journeys %
such that $d_E( \bar{Z}^{[a]} , \bar{Z}^{[b]} ) = 0 \implies \bar{Z}^{[a]} \equiv \bar{Z}^{[b]}$, and $d_E( \bar{Z}^{[a]} , \bar{Z}^{[b]} ) < d_E( \bar{Z}^{[a]} , \bar{Z}^{[c]} )$ implies that $\bar{Z}^{[a]}$ is more similar to $\bar{Z}^{[b]}$ than to $\bar{Z}^{[c]}$. %
On a high level, this metric conveys the \emph{accumulated} separation (or distance) between individual steps of two normalised counterfactual journeys. %
Among others, such a formulation allows to distinguish (and quantify the divergence of) paths %
that take different routes to arrive at the same counterfactual instance. %
In this case, the (user-specified) weight vector $w$ can prioritise journeys that stay close to each other at their beginning and only separate farther down the line. %

\begin{figure}[t]
    \centering
    \includegraphics[width=0.330\textwidth]{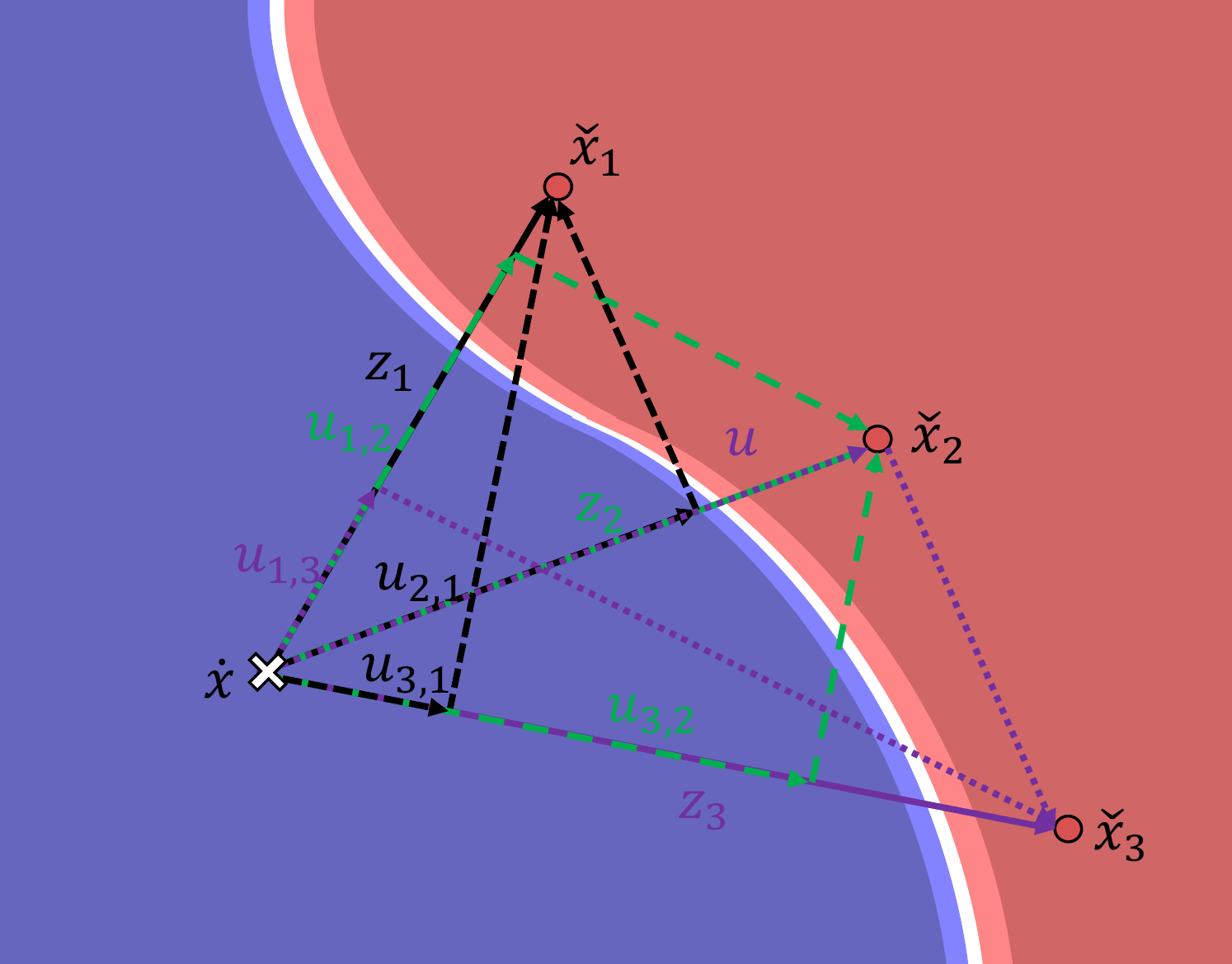}%
    \hspace{2em}%
    \scriptsize
    \begin{tabular}[b]{@{}r@{\hskip 5pt}r@{\hskip 3pt}r@{\hskip 3pt}r@{\hskip 3pt}r@{}}
    \toprule
    \multirow{3}{*}{\rotatebox[origin=c]{90}{\parbox[c]{1.35cm}{\vfill{}~compare to\vfill{}}}}
    & \multicolumn{4}{c}{reference path} \\
    & & $\check{x}_1$ & $\check{x}_2$ & $\check{x}_3$ \\
    & $\check{x}_1$ & 1.00 & 0.75 & 0.20 \\
    & $\check{x}_2$ & 0.90 & 1.00 & 0.65 \\
    & $\check{x}_3$ & 0.35 & 1.00 & 1.00 \\
    \cline{3-5}
    & & 0.75 & \textbf{0.92} & 0.62 \\
    \bottomrule
    \end{tabular}
    \caption{%
    Visual depiction of calculating \emph{opportunity potential} (left) and its example values (right). %
    Each element $l_{a,b}$ of the metric table conveys how far along the reference counterfactual path from $\mathring{x}$ to $\check{x}_a$ we can travel while still getting closer -- albeit in a possibly sub-optimal way -- to another path between $\mathring{x}$ and $\check{x}_b$; %
    the numbers at the bottom of the table (best in bold) capture the overall opportunity potential of a path in relation to all the other paths under consideration. %
    In this case the paths are assumed to be standalone direct vectors between the factual data point and counterfactual instances. %
    For example, if travelling along $z_1$ from $\mathring{x}$ to $\check{x}_1$, we can move toward the target ($\check{x}_1$) and at the same time get closer to $\check{x}_2$; %
    on the other hand, travelling the same path allows us to only contribute to a small fraction of the $\check{x}_3$ path, thus $l_{1,2} > l_{1,3}$. %
    }%
    \label{fig:choicetradeoff}%
\end{figure}

\paragraph{Measuring Opportunity Potential}%
To measure how much following a path towards one counterfactual contributes to reaching another counterfactual point we propose a novel spatially-aware metric called \emph{opportunity potential} and denoted by $l_{a,b}$. %
It answers the question: ``How close can I get from $\mathring{x}$ to $\check{x}_a$ (reference path) while at the same time getting closer to $\check{x}_b$ (comparison path)?'' %
To this end, the metric quantifies the fraction of the reference path $z_a$ -- which we assume here to be an optimal path between $\mathring{x}$ and $\check{x}_a$ -- that contributes to reaching the comparison path $z_b$; this process is demonstrated in Figure~\ref{fig:choicetradeoff}. %
Assuming a continuous metric space, the optimal path between a factual point $\mathring{x}$ and a counterfactual instance $\check{x}_a$ is given by the direct (shortest) vector linking the two: $z_a = \check{x}_a - \mathring{x}$. %
To compute the metric we first identify the vector orthogonal to $z_a$ that intersects with $\check{x}_b$. We find the point of intersection between this vector and $z_a$, and denote the difference between the intersection and the factual as $u_{a,b}$. The metric is therefore %
\begin{equation*}
    l_{a,b} = \min\left(\frac{\lVert u_{a,b} \rVert}{\lVert z_a \rVert}, \; 1\right)
    \text{~.}
\end{equation*}

To locate the intersection point, we parameterise the line between $\mathring{x}$ and $\check{x}_a$ as
\begin{equation*}
    g(l) = (1-l)\mathring{x} + l\check{x}_a - \check{x}_b
    \text{~,}
\end{equation*}
and find the closest point along this line to $\check{x}_b$, i.e., %
the value of $l$ that minimises $\lVert g(l) \rVert^2$, which is given by $l=-\frac{z_a \cdot z_b}{z_a \cdot z_a}$ (the proof is given in Appendix~\ref{apx:cose_point}). %
Therefore, we can reformulate the metric as %
\begin{equation*}
    l_{a,b} = 
    \begin{cases}
        0 &\text{if} \; l < 0 \\
        l &\text{if} \; l \in  [0,1] \\
        1 &\text{otherwise.}
    \end{cases}
\end{equation*}
The metric calculation process is formalised in Algorithm~\ref{alg:vector-metric}. %
If the optimal path is composed of a \emph{series} of vectors $Z^{[a]}$ as opposed to a single vector, we can apply the metric calculation procedure intended for graphs, which is outlined in Section~\ref{sec:graph}, instead of Algorithm~\ref{alg:vector-metric}. %
While the formulation of opportunity potential is blind to journeys that take different routes to arrive at the same counterfactual instance, the \emph{direction difference} metric described earlier accounts for such scenarios. %

\begin{algorithm}[t]
\caption{%
Calculate opportunity potential given by the shared proportion of \emph{direct} counterfactual vectors.%
}%
\label{alg:vector-metric}%
\begin{algorithmic}[1]
    \REQUIRE
        factual instance $\mathring{x}$;
        reference counterfactual point $\check{x}_a$;
        comparison counterfactual point $\check{x}_b$. 
    \ENSURE
        opportunity potential $l_{a,b}$
        of implementing $\check{x}_a$ as a direct, i.e., optimal, counterfactual vector %
        with respect to $\check{x}_b$. %
    \STATE
    $z_a \gets \check{x}_a - \mathring{x}$
    \COMMENT{direct vector between factual and counterfactual instances} %
    \STATE
    $z_b \gets \check{x}_b - \mathring{x}$
    \STATE $l \gets -\frac{z_a \cdot z_b}{z_a \cdot z_a}$
    \IF {$l < 0$}
        \STATE $l_{a,b} \gets 0$
    \ELSIF{$0 < l < 1$}
        \STATE $l_{a,b} \gets l$
    \ELSE
        \STATE $l_{a,b} \gets 1$
    \ENDIF
\end{algorithmic}
\end{algorithm}

\section{Implementing Explanatory Multiverse with Directed Graphs\label{sec:graph}}%

Applying changes to one's situation to achieve the desired outcome is an inherently continuous and incremental process~\cite{barocas2020hidden}. %
Given the extended period of time that it can span, this process may fail or be abandoned, e.g., due to an unexpected change in circumstances that invalidates the original plan, hence it may require devising alternative routes to the envisaged outcome. %
However, some of the actions taken thus far may make the adaptation to the new situation difficult or even impossible, prompting us to consider the loss of opportunity resulting from counterfactual path branching. %
To this end, explanatory multiverse extends the concept of counterfactual paths by accounting for uni-directional changes of features that reflect their monotonicity (as well as immutability) and recognising feature tweaks that cannot be undone. %

To demonstrate real-life benefits of such a comprehensive approach %
we propose an %
efficient and versatile implementation of explanatory multiverse based on \emph{directed graphs} -- see Algorithm~\ref{alg:graph}. %
Specifically, it %
builds upon FACE~\cite{poyiadzi2020face} %
by extending its $k$-nearest neighbours ($k$-NN) approach, which generates counterfactuals with the shortest path algorithm~\cite{dijkstra2022note}. %
The practicality of this implementation, nonetheless, comes at the expense of %
its somewhat imperfect application of the %
theoretical foundations laid down by the vector spaces conceptualisation. %

\begin{algorithm}[b]%
\caption{%
Build graph-based explanatory multiverse.%
}%
\label{alg:graph}%
\begin{algorithmic}[1]
    \REQUIRE
        data set $X$;
        probabilistic model $\widetilde{f}$; %
        instance to be explained $\mathring{x}$; %
        neighbours number $k$; %
        counterfactual classification threshold $t$. %
    \ENSURE 
        directed data graph $G = (V, E)$; %
        candidate counterfactuals $\check{X}$; %
        branching factor of paths $R$. \\ %
    \FOR[form graph edges]{every pair $(x_i, x_j) \in X$}%
        \STATE $e_{i,j} \gets d(x_i, x_j)$ %
        \COMMENT{compute distance}
    \ENDFOR
    \STATE $V \gets X$\\
    \STATE $E \gets \mathit{prune}$(E, k)
    \COMMENT{build neighbourhood graph by keeping $k$-NN for each node}
    \STATE $x_i \equiv \mathring{x}$
    \FOR{$x_j \in X$}
        \IF{$\widetilde{f}(x_j) \geq t$}
            \STATE $\check{X} \gets \check{X} \cup x_j$
            \STATE $V_{i,j}$, $E_{i,j} \gets \mathit{dijkstra}(G, x_i, x_j)$ %
            \COMMENT{find shortest path and the corresponding vertices}
            \STATE $r_{i,j} \gets \mathit{branching}(G, E_{i,j})$ %
            \STATE $R \gets R \cup r_{i,j}$
        \ENDIF
    \ENDFOR
\end{algorithmic}
\end{algorithm}

\paragraph{Path Branching}
Let $r_i$ denote the \emph{branching factor} of a node $v_i$, and $r_{1, n}$ be the branching factor of a path $E_{1,n}$ between nodes $v_1$ and $v_n$ consisting of $n$ steps. %
$r_i$ can be defined as the \emph{average of the shortest distance} from a vertex $v_i$ to each accessible node of the counterfactual class, or of all alternative classes for multi-class classification; its formulation depends on the data set and problem domain (a specific example is provided later in this section). %
Since there may be an overwhelmingly large number of accessible counterfactual instances (nodes), we can reduce the number of candidate points by introducing diversity criteria, e.g., (absolute) feature value difference, thus consider only a few representative counterfactuals. %
$r_{1, n}$ is defined as the average of individual branching factors $r_i$ corresponding to vertices $v_i$ that are travelled through when following the edges $e_{i, i+1}$ of the counterfactual path $E_{1,n}$, and computed as %
$$
    r_{1,n} = \frac{1}{n-2} \sum_{i=1}^{n-1} r_i
    \text{~.}
$$
The first and last nodes are excluded since the former is shared among all the paths and we stop exploring beyond the latter. %

To account for choice complexity, we can introduce a discount factor $\gamma > 0$ that allows us to reward ($\gamma > 1$) or penalise ($\gamma < 1$) branching in the early stages of a path, i.e., %
$$
        r_{1,n}
        =
        \frac{1}{n-2} \sum_{i=1}^{n-1} \gamma^{i-1} r_i
         \text{~.}
$$
The choice of $\gamma$ should be informed by background knowledge and domain expertise; it can also reflect user preferences. %
For example, when generating counterfactual paths in the medical domain, rewarding early branching ($\gamma>1$) allows to discover various initial treatment options %
to better tailor the disease management strategy to the patient's needs and wants. %
Late branching ($\gamma<1$), on the other hand, enables exploring how a specific treatment plan can be adjusted in the future, e.g., depending on the patient's response to medications. %
Tweaking $\gamma$ can therefore help to address the unique requirements and expectations of diverse explainees. %

\paragraph{Constraining Feature Changes}%
Our directed graph implementation accounts for feature monotonicity to capture uni-directional changes of relevant attribute values. %
For some data domains, however, imposing such a strict assumption may be too restrictive and yield no viable explanations, as is the case in our next example -- path-based counterfactuals for the MNIST data set of handwritten digits. %
Here, the explanations capture (step-by-step) transitions between different digits, a process that is realised through \emph{addition} of individual pixels; %
the plausibility of such paths is enforced by composing them exclusively of instances that were observed before (and stored in a dedicated data set). %

As noted above, such a restrictive setup yields no viable counterfactual paths, which prompts us to %
relax the feature monotonicity constraint by allowing pixels to also be removed; however, we specify this action to be $\lambda$ times more costly than adding pixels, %
where $\lambda > 1$ is a penalty term discouraging feature changes in the undesirable direction.%
\footnote{When $\lambda=1$, the formulation becomes equivalent to that of unweighted $L^2$ distance. %
When $\lambda<1$, removing pixels becomes more desirable than adding them; %
this is equivalent to setting $\lambda>1$ for the task of adding pixels.} %
This formalisation simulates scenarios where backtracking steps of a counterfactual path is undesirable, which in the case of MNIST can be viewed as removing pixels that are already in an image (whether pre-existing or added during recourse). %
The corresponding distance between two nodes $v_a$ and $v_b$ can be defined as %
\begin{equation}\label{eq:weight_dist}%
    d_{\lambda}(v_a, v_b) = \sqrt{\sum_{i=1}^m \left(
    \phi_\lambda(v_{a, i}, v_{b, i})
    (v_{a, i} - v_{b, i})
    \right)^{2}}\text{~,}%
\end{equation}%
where %
$v_{a,i}$ is the $i$\textsuperscript{th} ($1 \leq i \leq m$) feature of the instance $x_a$ represented by the node $v_a$ and %
$$
    \phi_\lambda(v_{a, i}, v_{b, i}) =
    \begin{cases}
        -\lambda &\quad \text{if} \;  v_{a, i} - v_{b, i} < 0\\
        1 &\quad \text{otherwise.} %
    \end{cases}
$$

\begin{figure}[t]
    \centering
    \includegraphics[trim={145pt 5pt 10pt 0pt},clip,width=.60\linewidth]{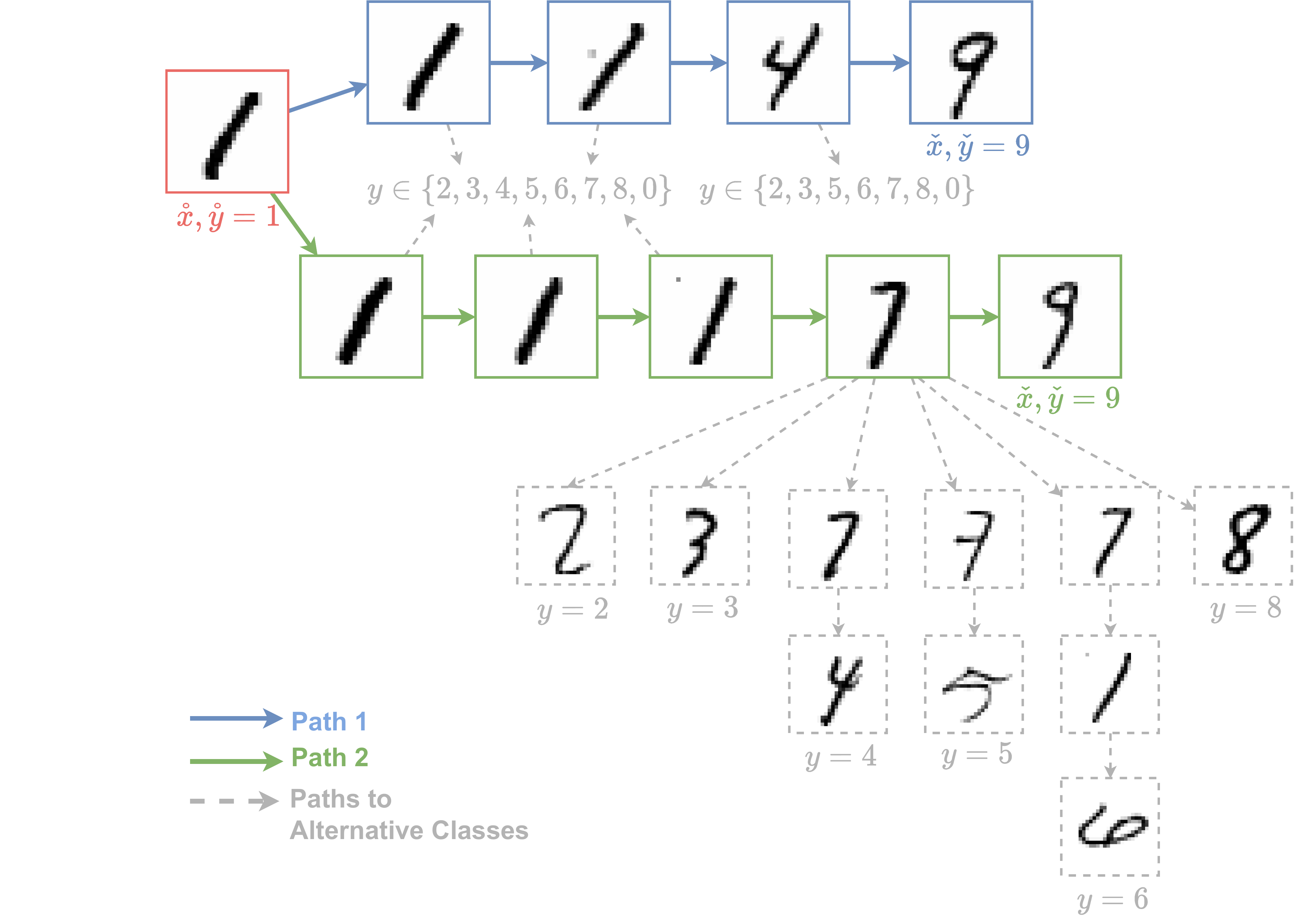}%
    \caption{%
    Example counterfactual journeys identified in the MNIST data set of handwritten digits. %
    Paths~1 (blue) and~2 (green) are counterfactual explanations for an instance $\mathring{x}$ classified as digit~$1$ ($\mathring{y} = 1$) with the desired counterfactual class being digit~$9$ ($\check{y} = 9$). %
    Paths leading to alternative counterfactual instances of classes other than~$9$, i.e., $y \in \mathcal{Y} \setminus (1, 9)$, are also possible (shown in grey). %
    Path~1 is shorter than Path~2 at the expense of explainees' agency, which is reflected by the \emph{branching factor} score of 0.38 versus 0.41; %
    therefore, switching to alternative paths that lead to different classes while travelling along Path~1 is more difficult, i.e., more costly in terms of distance. %
    }%
    \label{fig:mnist}%
\end{figure}

In other application areas, feature constraints can additionally capture the mutability of attributes as well as the directionality (increase or decrease) of their change and the possible range thereof; %
specifying these properties, nonetheless, often requires individual user input or domain expertise. %
For example, in the German Credit data set, altering the amount of savings of a person can produce a valid counterfactual explanation, %
but achieving this goal may be problematic for some individuals, making such a solution unsatisfactory. %
To account for this, we can penalise the increase of a (selected) feature value by making it $\lambda > 1$ times more costly than the equivalent decrease. %
If the perceived difficulty of an attribute change is negligible, $\lambda$ can be set close to $1$ to reflect that; %
otherwise, it should be proportionally larger (or smaller) to capture the challenges associated with modifying the values of selected features. %
This brief example clearly demonstrates that %
there is no one-size-fits-all $\lambda$ setting %
and that this parameter needs to be determined %
on a case-by-case basis, accounting for domain-specific constraints and (individual) user preferences. %

Continuing with the MNIST example, we apply Algorithm~\ref{alg:graph} with $k=20$ using $d_{\lambda}$ as the distance function, setting $\lambda = 1.1$. %
Figure~\ref{fig:mnist} shows example counterfactual paths starting at $f(\mathring{x}) = 1$ and terminating at $f(\check{x}) = 9$. %
The branching factor of a node $v_i$ representing a digit image is computed as $r_i=-\log(c(v_i))$, where %
$$
    c(v_i) =
    \frac{1}{\lvert \mathcal{Y}^\prime \rvert}
    \sum_{y \in \mathcal{Y}^\prime}
    \sum_{e_{m,n} \in E_{i, z} }
    d_{\lambda} (v_m, v_n)
    \;\;\; \text{s.t.} \;\;\; %
    f(v_z) = y
    \text{~.}
$$
Here, %
$\mathcal{Y}^\prime \equiv \mathcal{Y} \setminus (\mathring{y}, \check{y})$, i.e., we exclude the factual and counterfactual classes; %
$E_{i, z}$ is the shortest path from $v_i$ to all nodes $v_z$ that represent alternative counterfactual instances, i.e., $f(v_z) \in \mathcal{Y}^\prime$. %
The path length indicates the number of pixels changed when transforming a factual data point into a counterfactual instance. %
Branching factor of a path can be interpreted as the ease of switching to alternative paths. %
A high branching factor for a node indicates more agency and ease of switching to counterfactual paths leading to alternative explanations. %
A branching factor that decreases after making a step signals that the current path begins to diverge from its viable alternatives. %

\paragraph{Measuring Opportunity Potential}%
When dealing with a (directed) graph $G=(V, E)$, travelling from the factual point $\mathring{x}$ to any counterfactual instance $\check{x}$ requires following the graph's edges. %
The \emph{opportunity potential} metric $l_{a,b}$ in this case quantifies how much traversing the \emph{optimal} path to a \emph{reference} counterfactual $\check{x}_a$ contributes to approaching another counterfactual $\check{x}_b$. %
Here, the optimal, or shortest, path is given by a collection of edges whose summed weights are the smallest -- as opposed to a direct vector connecting the factual and counterfactual instances in the feature space employed in the vector-based conceptualisation of explanatory multiverse (see Section~\ref{sec:vector}). %

Assume that $V_{1,a} \subseteq V$ and $E_{1,a} \subseteq E$ are respectively the sets of vertices and edges connecting $\mathring{x}$ to $\check{x}_a$ along the optimal path, and that $\lvert V_{1,a} \rvert$ denotes the number of vertices in this set, with $v_1 \equiv \mathring{x}$, $v_{\lvert V_{1,a} \rvert} \equiv \check{x}_a$ and $\mathit{sum}(E_{1,a})$ being the length of this path. %
Computing \emph{opportunity potential} then relies on traversing the vertices of the \emph{reference path} connecting $\mathring{x}$ to $\check{x}_a$ and checking at every step whether the length of the path from this point to the alternative counterfactual $\check{x}_b$ begins to increase. %
Such an observation signals that the reference path $\check{x}_a$ starts to diverge -- i.e., move away -- from $\check{x}_b$, which marks the point along it where the contribution to $\check{x}_b$ is the largest. %
This process is formalised in Algorithm~\ref{alg:graph-metric}. %

\begin{algorithm}[b]
\caption{%
Calculate opportunity potential given by the shared path edges in graph-based explanatory multiverse.%
}%
\label{alg:graph-metric}%
\begin{algorithmic}[1]
    \REQUIRE
        factual instance $\mathring{x}$;
        reference counterfactual point $\check{x}_a$;
        comparison counterfactual point $\check{x}_b$;
        directed data graph $G = (V, E)$ from Algorithm~\ref{alg:graph}. %
    \ENSURE
        opportunity potential $l_{a,b}$
        of implementing $\check{x}_a$ via its shortest, i.e., optimal, counterfactual path %
        with respect to $\check{x}_b$. %
    \STATE $V_{a}$, $E_{a} \gets \mathit{dijkstra}(G, \mathring{x}, \check{x}_a)$ %
            \COMMENT{find shortest (optimal) path and the corresponding vertices}
    \STATE $l \gets 0$
    \FOR{$i \gets 2$; \quad $i \gets i + 1$; \quad $i \leq \lvert V_a \rvert$}%
    \STATE $x_i \equiv v_i$
    \COMMENT{note $v_1 \equiv x_1 \equiv \mathring{x}$ and $v_{|V_a|} \equiv x_{|V_a|} \equiv \check{x}_a$}%
    \STATE $e_i \gets \mathit{edge\_connecting}(v_{i-1}, v_i)$
    \STATE $V_{i}$, $E_{i} \gets \mathit{dijkstra}(G, x_i, \check{x}_b)$ %
    \IF{$\mathit{sum}(E_i) \geq \mathit{sum}(E_{i-1})$}
        \STATE break
    \ELSE
        \STATE $l \gets l + e_i$
    \ENDIF
    \ENDFOR
    \STATE $l_{a,b} \gets \frac{l}{\mathit{sum}(E_a)}$
\end{algorithmic}
\end{algorithm}

\begin{figure}[t]%
    \centering
    \includegraphics[width=0.45\textwidth]{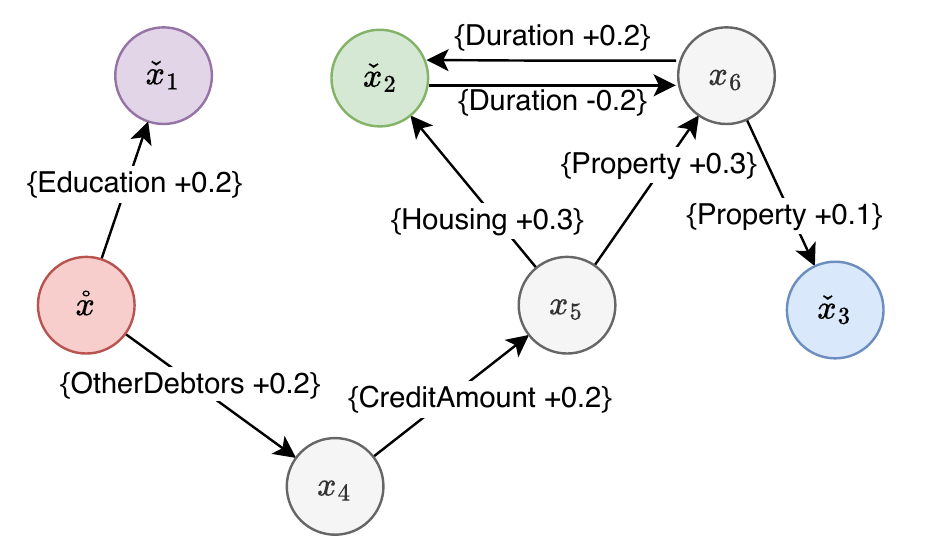}%
    \hspace{2em}%
    \scriptsize
    \begin{tabular}[b]{@{}r@{\hskip 5pt}r@{\hskip 3pt}c@{\hskip 3pt}c@{\hskip 3pt}c@{}}
    \toprule
    \multirow{3}{*}{\rotatebox[origin=c]{90}{\parbox[c]{1.35cm}{\vfill{}~compare to\vfill{}}}}
    & \multicolumn{4}{c}{reference path} \\
    & & $\check{x}_1$ & $\check{x}_2$ & $\check{x}_3$ \\
    & $\check{x}_1$ & 1 & 0 & 0 \\
    & $\check{x}_2$ & 0 & 1 & \nicefrac{7}{8} \\
    & $\check{x}_3$ & 0 & \nicefrac{4}{7} & 1 \\
    \cline{3-5}
    & & 0.33 & 0.52 & \textbf{0.63} \\
    \bottomrule
    \end{tabular}
    \caption{
    Visual representation of three counterfactuals -- $\check{x}_1$, $\check{x}_2$ and $\check{x}_3$ -- for the $\mathring{x}$ instance in a loan application scenario using the (directed) graph-based explanatory multiverse (left), %
    and the corresponding \emph{opportunity potential} values (right). %
    Each element $l_{a,b}$ of the metric table conveys how far (determined by the sum of edge weights) along the reference (shortest and optimal) counterfactual path between $\mathring{x}$ and $\check{x}_a$ we can travel while still getting closer -- albeit in a possibly sub-optimal way -- to another path between $\mathring{x}$ and $\check{x}_b$; %
    the numbers at the bottom of the table (best in bold) capture the overall opportunity potential of a path in relation to all the other paths under consideration. %
    Here, the (optimal) path from $\mathring{x}$ to $\check{x}_3$ maximises the explainee's opportunity to switch to alternative paths, which is captured by the highest opportunity potential of 0.63. %
    }%
    \label{fig:graphchoicetradeoff}%
\end{figure}

A demonstration of calculating opportunity potential in a loan application scenario is given in Figure~\ref{fig:graphchoicetradeoff}. %
This example is inspired by the German Credit data set where a user $\mathring{x}$ who is denied a loan, i.e., $\mathring{y} = 0$, can purse three different counterfactual paths -- to $\check{x}_1$, $\check{x}_2$ or $\check{x}_3$ -- to reverse this decision and receive the loan, i.e., $\check{y} = 1$. %
Here, an edge of the graph corresponds to an action that affects a particular feature value. %
By travelling along the shortest path from $\mathring{x}$ to $\check{x}_3$ -- given by the $ \mathring{x} \rightarrow x_4 \rightarrow x_5 \rightarrow x_6 \rightarrow \check{x}_3 $ vertices -- we can reach the target and at the same time approach $\check{x}_2$ via the $ \mathring{x} \rightarrow x_4 \rightarrow x_5$ section of this path. %
On the other hand, while the path to $\check{x}_1$ offers the smallest distance -- since only one short step $ \mathring{x} \rightarrow \check{x}_1 $ is required -- it does not contribute to reaching alternative counterfactuals. %
In our example, the path to $\check{x}_3$ has the highest and the path to $\check{x}_1$ has the lowest opportunity potential. %

Another (graphical) illustration of quantifying opportunity potential for counterfactual paths, this time using the MNIST data set, can be found in Appendix~\ref{apx:examples}. %
In both examples, we can see a clear trade-off between opportunity potential and explanation length (which can be thought of as its cost). %
Our comprehensive experimental results -- reported in the following section -- %
validate this observation. %

\section{Experimental Evaluation\label{sec:experiment}}%

To further ground our contributions, we employ the graph-based explanatory multiverse implementation -- which we call \textsc{FACElift} -- %
given by Algorithms~\ref{alg:graph} and~\ref{alg:graph-metric} %
to generate a diverse range of counterfactual paths while taking into account their spatial relation. %
Specifically, we first apply it to %
three tabular binary classification data sets: %
German Credit, Adult Income and Credit Default (Section~\ref{sec:german-results}); %
next, we run it on %
three multi-class image classification data sets: %
MNIST handwritten digits %
(Section~\ref{sec:mnist-results}) %
as well as BreastMNIST and PneumoniaMNIST from the MedMNIST medical imaging collection (Section~\ref{sec:med-mnist-results}). %
We compare our method against other state-of-the-art explainers in terms of counterfactual \emph{distance} %
and \emph{opportunity potential}. %
We finish this section with an ablation study (Section~\ref{sec:experiment:ablation}). %

\subsection{German Credit, Adult Income and Credit Default (Tabular Data)\label{sec:german-results}}%

We first experiment with three tabular data sets -- namely, German Credit, Adult Income and Credit Default -- to demonstrate increased \emph{opportunity potential} offered by \textsc{FACElift}. %

\paragraph{Data Sets and Classifiers}%
The German Credit data set consists of $m=20$ categorical and numerical features; %
the binary target variable encodes either a \emph{good} or a \emph{bad} credit rating, denoted by $y = 1$ and $y = 0$ respectively. %
The Adult Income data set contains $m = 14$ attributes; %
the target variable denotes \emph{high}, with $y=1$, or \emph{low}, with $y=0$, income. %
The Credit Default data set includes $m=23$ features; %
the target variable indicates whether an individual \emph{has}, with $y=0$, or \emph{has not}, with $y=1$, previously \emph{defaulted} on payments. %
For all the three data sets, $y=1$ is the more desirable class. %
Following a common practice in explainability studies~\cite{pawelczyk2022probabilistic}, we one-hot encode categorical attributes, which allows us to treat all the input features as continuous variables, and normalise the remaining attributes to the $[0,1]$ range, i.e., $\mathcal{X} = [0,1]^m$. %
Using 80:20 ratio we then split the data sets into two parts that we use respectively for model training and evaluation. %
For each data set we train a simple neural network with a 50-neuron hidden layer and ReLU activation function. %
Given the small size of the German Credit data set, we generate recourse for all its 257 data points (sourced from both its training and test set) predicted with \emph{bad} credit rating. %
For the Adult Income and Credit Default data sets, we generate recourse for instances predicted with the undesirable class found exclusively in the test sets. %

\paragraph{Counterfactual Explainers}
We configure %
\textsc{FACElift} %
such that it %
constructs the neighbourhood graph by connecting each instance to its $k=20$ nearest neighbours. %
To ensure universality of the resulting data representation, %
the distance between any two points is calculated %
without imposing any feature change constraints, %
therefore %
$\lambda$ is set to $1$ in Equation~\ref{eq:weight_dist}. %
For each factual instance, we configure \textsc{FACElift} to identify its top $c$ closest counterfactuals in the neighbourhood graph; %
we use $c \in \{5, 10\}$, with the impact of this parameter choice discussed later in our ablation study (Section~\ref{sec:experiment:ablation}). %
Using Algorithm~\ref{alg:graph-metric}, %
we then calculate opportunity potential %
for each of the top $c$ closest counterfactuals %
and select the one with the largest value of this metric as the optimal explanation. %

We choose three baseline explainers that can generate multiple counterfactuals for a single instance. %
FACE provides a sequence of existing data instances as a path to the target counterfactual based on a neighbourhood graph~\cite{poyiadzi2020face}. %
GrowingSphere is a search-based explainer that gradually expands its search range until it finds a counterfactual instance~\cite{laugel2023achieving}. %
Prototype is an explainer that encodes the data set into a latent space from which it then retrieves counterfactual examples that are close to prototypical instances of the target class~\cite{van2021interpretable}. %
We set $k=20$ for FACE to ensure its competitiveness with \textsc{FACElift}; %
we also perform appropriate hyperparameter tuning for the other two baselines to guarantee their optimal performance. %
While all of our baseline explainers strive to identify the least distant feasible counterfactual instance, %
\textsc{FACElift} outputs the counterfactual path with the highest opportunity potential among top $c$ closest explanations. %

Since our work focuses on spatially-aware desiderata of counterfactual paths, for each factual data point we require both the counterfactual instance and a path leading to it. %
While FACE outputs this information by design, %
GrowingSphere and Prototype are not capable of it. %
To adapt these explainers to our experiments we employ a post-hoc path generator for counterfactual instances that is based on Binary Space Partitioning and formalised by Algorithm~\ref{alg:bsp} given in Appendix~\ref{app:bsp}. %

\paragraph{Evaluation}
We use the $L^2$-norm to calculate the \emph{distance} between the factual and counterfactual instances; %
this measure captures %
the \emph{cost} of implementing the explanation. %
We also report %
the \emph{average} of %
\emph{opportunity potential} %
computed %
between the \emph{optimal} counterfactual and a set of its \emph{five} (sub-optimal) alternatives generated by an explainer. %
For the baselines, the best counterfactual is that closest to the factual instance, with its alternatives being increasingly farther away. %
The latter set is produced by a state-of-the-art method designed to %
output %
the nearest counterfactuals that are (at least) one unit of distance apart from each other, %
hence ensuring their diversity and coverage~\citep{karimi2020model}. %
For \textsc{FACElift}, the optimal explanation is determined based on both its proximity and opportunity potential -- as described earlier -- %
and the alternatives are simply counterfactuals that are ranked lower on these criteria. %

\begin{table*}[t]
\centering
\setlength{\tabcolsep}{3.5pt}
    \footnotesize
\begin{tabular}{@{}l rr rrrr@{}}
\toprule
 & \multicolumn{2}{c}{German Credit} & \multicolumn{2}{c}{Adult Income} & \multicolumn{2}{c}{Credit Default}  \\
 \cmidrule(lr){2-3}\cmidrule(lr){4-5}\cmidrule(lr){6-7}
 & distance $\downarrow$ & opportunity $\uparrow$ & distance $\downarrow$ & opportunity $\uparrow$ & distance $\downarrow$ & opportunity $\uparrow$ \\ 
 \midrule
FACE & 2.16 $\pm$ 0.72 & 0.31 $\pm$ 0.30 & 3.59 $\pm$ 0.44 & 0.30 $\pm$ 0.18 & 1.10 $\pm$ 0.29 & 0.21 $\pm$ 0.16 \\
GrowingSphere & 1.68 $\pm$ 1.49 & 0.36 $\pm$ 0.22 & \textbf{2.64} $\pm$ 0.97 & 0.38 $\pm$ 0.14 & \textbf{0.58} $\pm$ 0.31 & 0.27 $\pm$ 0.21 \\
Prototype & \textbf{1.14} $\pm$ 1.11 & 0.32 $\pm$ 0.15 & 2.84 $\pm$ 0.52 & 0.27 $\pm$ 0.18 & 0.74 $\pm$ 0.65 & 0.32 $\pm$ 0.24 \\
\textsc{FACElift} ($c=5$) & 2.40 $\pm$ 0.99 & 0.61 $\pm$ 0.11 & 3.71 $\pm$ 0.87 & 0.82 $\pm$ 0.16 & 1.28 $\pm$ 0.53 & 0.73 $\pm$ 0.18 \\ 
\textsc{FACElift} ($c=10$) & 2.54 $\pm$ 0.99 & \textbf{0.65} $\pm$ 0.10 & 3.90 $\pm$ 0.35 & \textbf{0.88} $\pm$ 0.32 & 1.34 $\pm$ 0.41 & \textbf{0.79} $\pm$ 0.12 \\
\bottomrule
\end{tabular}
\caption{%
Comparison of explainers for three tabular data sets: German Credit, Adult Income and Credit Default. %
Recourse \emph{distance} is measured with the $L^2$-norm; %
\emph{opportunity} potential quantifies the contribution of implementing the most optimal explanation to reaching its alternatives %
(the calculation of this metric is explained in Section~\ref{sec:graph}, the best result for each metric is displayed in bold). %
\textsc{FACElift} consistently delivers the best opportunity potential while only incurring a small distance penalty -- which is an inherent trade-off of these two desiderata -- across all the data sets. %
}
\label{tab:tabular-results}
\end{table*}

\paragraph{Results}
Table~\ref{tab:tabular-results} reports our results. %
For the German Credit data set, the Prototype explainer generates the closest counterfactuals, but their opportunity potential is poor. %
Similarly, it maintains relatively low distance on Adult Income and Credit Default, but the opportunity potential remains low. %
The explanations output by GrowingSphere %
are the closest %
for Adult Income and Credit Default, %
nonetheless their opportunity potential leaves room for improvement. %
Across all the three data sets, %
FACE offers counterfactuals that are farther from the factual instance and, in most cases, with the worst opportunity potential; %
this is likely the price to pay for guaranteeing a feasible path connecting the two. %

Among the baselines, either GrowingSphere or Prototype achieves the shortest distances to the counterfactual, but their highest opportunity potential is significantly ($p<0.001$) lower than that of \textsc{FACElift} for $c\in\{5,10\}$. %
Our method finds explanations with the best opportunity potential at the expense of their increased distance, which is expected given the inherent trade-off between the two desiderata. %
This compromise becomes more apparent when changing $c$ from $5$ to $10$, which enables \textsc{FACElift} to find counterfactuals with better opportunity potential by exploring more distant explanations (i.e., those with longer paths). %
Crucially, independent $t$-tests reveal statistically significant differences in both distance and opportunity potential between \textsc{FACElift} and our baselines. %
These results confirm the pertinence of the aforementioned trade-off and highlight \textsc{FACElift}'s overall superior performance. %

\subsection{MNIST (Image Data)\label{sec:mnist-results}}%

We chose MNIST for our second set of experiments as this data set facilitates: %
\begin{enumerate*}[label=(\arabic*)]
\item
appealing and intuitive visualisations of spatially intertwined counterfactual paths as well as their direct comparison to the corresponding explanations output by FACE~\cite{poyiadzi2020face}; %
\item
illustration of the explanatory multiverse's inherent capability to deal with multi-class classification -- a topic that is rarely explored in explainability literature~\cite{sokol2020limetree}; and %
\item
demonstration of \textsc{FACElift}'s ability to deal with diverse data types. %
\end{enumerate*}

\paragraph{Data Set and Classifier}%
The data set consists of handwritten digits from 0 to 9, which are the class labels. %
Each image is 28$\times$28 pixels, and each pixel takes a value between 0 and 255, i.e., the images are greyscale. %
Following common practice we normalise the feature range, i.e., pixel intensity, to the $[0, 1]$ interval, i.e., $\mathcal{X} = [0, 1]^m$ where $m = 28 \times 28 = 784$~\cite{poyiadzi2020face,van2021interpretable}. %
For our experiments we randomly select 1,000 images of every digit, which yields 10,000 instances in total. %
Using 80:20 ratio we then split this subset into two parts that we use respectively for training and evaluation; %
our image classification model is a fully connected neural network with a 100-neuron hidden layer and ReLU activation function. %
To generate counterfactual explanations, we randomly select 100 data points that represent digit~1 from the test set. %
We use these as our factual instances, targeting all the other classes when generating counterfactual explanations. %

\paragraph{Counterfactual Explainers and Evaluation}
We set up \textsc{FACElift} as described in Section~\ref{sec:graph}, using $k=20$ for the neighbourhood graph and $\lambda=1.1$ for the weighted distance function. %
We experiment with two distinct scenarios when assessing the spatial relation between multiple counterfactual instances. %
First, we assume that the most desirable counterfactual instances as well as the alternative counterfactuals are all of the same class -- digit~9 in our experiments. %
This scenario captures the most prevalent counterfactual search task where the classification outcome is binary, and one result (class~9) is strictly better than the other (class~1). %
We further consider a largely neglected, multi-class explainability scenario where in addition to the most desirable counterfactual class we also permit a collection of alternative counterfactual classes~\cite{sokol2020limetree}. %
In our experiments, we search for counterfactuals of class~9 for a factual instance of class~1, but we also allow a transition to the remaining counterfactual classes \{0, 2, 3, 4, 5, 6, 7, 8\}. %
Illustrative examples of these two explainability scenarios are given in Figure~\ref{fig:mnist_calculation} placed in Appendix~\ref{apx:examples}. %

Our evaluation protocol is identical to the one %
outlined in Section~\ref{sec:german-results}. %
Here, distance is computed using Equation~\ref{eq:weight_dist} with $\lambda=1.1$. %
Also, when generating the set of five alternative counterfactuals for opportunity potential calculation, we require them to be \emph{five} units of distance apart from each other -- rather than just one -- to ensure their diversity. %

\begin{table*}[t]
\centering
\setlength{\tabcolsep}{3.5pt}
    \footnotesize
\begin{tabular}{@{}l rrrr rr rr@{}}
\toprule
 & \multicolumn{4}{c}{MNIST} & \multicolumn{4}{c}{MedMNIST} %
 \\
 \cmidrule(lr){2-5}\cmidrule(lr){6-9}
 & \multicolumn{2}{c}{$\mathring{y} = 1,\; \check{y}^{\prime} \in \{0,2,3,4,5,6,7,8\}$} & \multicolumn{2}{c}{$\mathring{y} = 1,\; \check{y}=9$} & \multicolumn{2}{c}{PneumoniaMNIST} & \multicolumn{2}{c}{BreastMNIST} \\
 \cmidrule(lr){2-3}\cmidrule(lr){4-5}\cmidrule(lr){6-7}\cmidrule(lr){8-9}
 & distance $\downarrow$ & opportunity $\uparrow$ & distance $\downarrow$ & opportunity $\uparrow$ & distance $\downarrow$ & opportunity $\uparrow$ & distance $\downarrow$ & opportunity $\uparrow$ \\ 
 \midrule
FACE & 79.33 $\pm$ 16.19 & 0.23 $\pm$ 0.10 & 79.33 $\pm$ 16.19 & 0.38 $\pm$ 0.19
& 23.15 $\pm$ 8.02 & 0.14 $\pm$ 0.06 & 28.58 $\pm$ 9.41 & 0.31 $\pm$ 0.07 \\
GrowingSphere & \textbf{71.66} $\pm$ 34.11 & 0.25 $\pm$ 0.20 & \textbf{71.66} $\pm$ 34.11 & 0.48 $\pm$ 0.34 & \textbf{21.38} $\pm$ 4.30 & 0.17 $\pm$ 0.05 & 26.36 $\pm$ 5.29 & 0.27 $\pm$ 0.14 \\
Prototype & 91.95 $\pm$ 47.92 & 0.20 $\pm$ 0.14 &  91.95 $\pm$ 47.92 & 0.53 $\pm$ 0.31 & 24.60 $\pm$ 5.22 & 0.21 $\pm$ 0.14 & \textbf{25.34} $\pm$ 7.21 & 0.15 $\pm$ 0.11 \\
\textsc{FACElift} ($c=5$) & 103.95 $\pm$ 16.61 & 0.35 $\pm$ 0.09 & 105.97 $\pm$ 21.05 & 0.59 $\pm$ 0.11 & 29.75 $\pm$ 7.77 & 0.51 $\pm$ 0.07 & 33.54 $\pm$ 8.85  & 0.57 $\pm$ 0.07 \\ 
\textsc{FACElift} ($c=10$) & 110.38 $\pm$ 17.48 & \textbf{0.36} $\pm$ 0.19 & 114.93 $\pm$ 22.73 & \textbf{0.63} $\pm$ 0.09 & 31.00 $\pm$ 6.78 & \textbf{0.58} $\pm$ 0.13 & 35.03 $\pm$ 8.08 & \textbf{0.64} $\pm$ 0.19 \\
\bottomrule
\end{tabular}
\caption{%
Comparison of explainers for the MNIST and MedMNIST -- specifically, its PneumoniaMNIST and BreastMNIST components -- data sets. %
\emph{Distance} is measured with $d_{\lambda}$ for the former and the $L^2$-norm for the latter; %
\emph{opportunity potential} is calculated as outlined in Section~\ref{sec:graph} -- see the caption of Table~\ref{tab:tabular-results} for more details (the best result for each metric is displayed in bold). %
In MNIST, the factual points represent digit~1, and digit~9 is chosen to be the most desirable counterfactual class. %
The first group of MNIST results (left) reports statistics for when the alternative counterfactual instances are required to be of a class other than digit~9. %
The second group of MNIST results (right) reports statistics for when the search is restricted to counterfactuals representing digit~9. %
In MedMNIST, both the target and alternatives are of class ``normal''. %
While \textsc{FACElift} yields longer counterfactual paths on average, it offers the highest opportunity potential, thus providing explainees with more agency. %
}
\label{tab:image-results}
\end{table*}

\paragraph{Results}
Table~\ref{tab:image-results} reports our results. %
In the multi-class explainability scenario opportunity potential of the most optimal path is overall lower than in the binary explainability setting. %
This is because alternative counterfactuals are more similar and closer to the optimal explanation when they are all of the same class. %
GrowingSphere generates the closest counterfactual explanations across the board, but it offers relatively low opportunity potential, %
whereas Prototype appears to balance these two desiderata; %
FACE yields mixed results as well. %
\textsc{FACElift} exhibits a similar trade-off: it achieves the highest opportunity potential at the expense of elongated paths. %
Its behaviour when increasing $c$ form $5$ to $10$ is similar to that observed in the tabular experiments (Section~\ref{sec:german-results}). %
Statistical testing in the multi-class setting proves that \textsc{FACElift}'s opportunity potential for $c \in \{5, 10\}$ is \emph{significantly} ($p<0.001$) higher than that offered by any of our baselines. %
This result also holds %
in the binary classification scenario (with $p<0.05$ for $c=5$ and $p<0.001$ for $c=10$). %
Our results on MNIST therefore align with those for tabular data, further demonstrating that \textsc{FACElift} consistently provides high user agency regardless of the data domain and application area. %

\subsection{BreastMNIST and PneumoniaMNIST (MedMNIST Image Data)\label{sec:med-mnist-results}}%

Next, we evaluate \textsc{FACElift} on two data sets -- BreastMNIST and PneumoniaMNIST -- from the MedMNIST collection. %
These experiments allow us to assess how well our method can deal with more complex, real-life image data coming from diverse (medical) application domains. %

\paragraph{Data Set and Classifier}
MedMNIST is a large-scale MNIST-like collection of standardised two-dimensional biomedical images that includes 12 separate data sets. %
To showcase \textsc{FACElift}'s ability to handle complex data of varying size, we select two %
of them for our experiments%
: the relatively small BreastMNIST data set, which consists of 780 breast ultrasound images; and the medium-sized PneumoniaMNIST, which contains 5,856 chest X-ray images. %
All the MedMNIST samples share the same properties as the MNIST images, therefore our pre-processing and train--test data spilt %
follow %
the procedures described in Section~\ref{sec:mnist-results}. %

For each data set %
we trained %
a neural network classifier based on the ResNet-18 architecture, %
using grid search for hyperparameter tuning; %
the resulting models achieved 80.7\% and 85.3\% accuracy respectively for BreastMNIST and PneumoniaMNIST. %
Since both data sets have a binary target variable, %
we specify the undesirable class to be ``Malignant'' ($y=0$) and the preferred class to be ``Normal'' ($y=1$) in BreastMNIST; %
similarly, we use ``Pneumonia'' as the negative ($y=0$) and ``Normal'' as the positive ($y=1$) class in PneumoniaMNIST. %
During evaluation, we generate counterfactuals for all the test set samples classified with the undesirable class ($y=0$). %
This gives us 42 instances from BreastMNIST and 234 data points from PneumoniaMNIST. %

\paragraph{Counterfactual Explainers and Evaluation}
Since the direction of feature value change has no semantic meaning in the context of these two data sets, %
we do not impose any constraints on attribute value modification, i.e., we set $\lambda=1$. %
Additionally, we set $k=10$ when constructing the neighbour graph for both FACE and \textsc{FACElift}. %
The evaluation itself follows the procedure used with the MNIST data set (Section~\ref{sec:mnist-results}) %
with the only difference being the distance function: here we employ the $L^2$-norm. %

\paragraph{Results}
Table~\ref{tab:image-results} reports our results, %
which lead to conclusions that are similar to those drawn for the %
tabular (Section~\ref{sec:german-results}) and MNIST (Section~\ref{sec:mnist-results}) data sets. %
Specifically, we find that either GrowingSphere or Prototype generates the closest counterfactuals, %
nonetheless both of these methods fail to achieve high opportunity potential, which is significantly lower ($p<0.001$) than that of \textsc{FACElift} explanations. %
While FACE produces longer paths, this increase in distance does not translate into substantial improvement in opportunity potential. %
\textsc{FACElift}, on the other hand, continues to achieve superior opportunity potential while incurring only a small distance penalty. %
Our extensive evaluation across six diverse tabular and image data sets %
of varying size and spanning diverse domains %
clearly shows %
that \textsc{FACElift} %
delivers %
counterfactuals with high opportunity potential while carefully managing their increased distance. %

\subsection{Ablation Study\label{sec:experiment:ablation}}
\begin{figure}[t]
    \centering
    \begin{subfigure}[b]{0.325\textwidth}%
        \includegraphics[width=\textwidth]{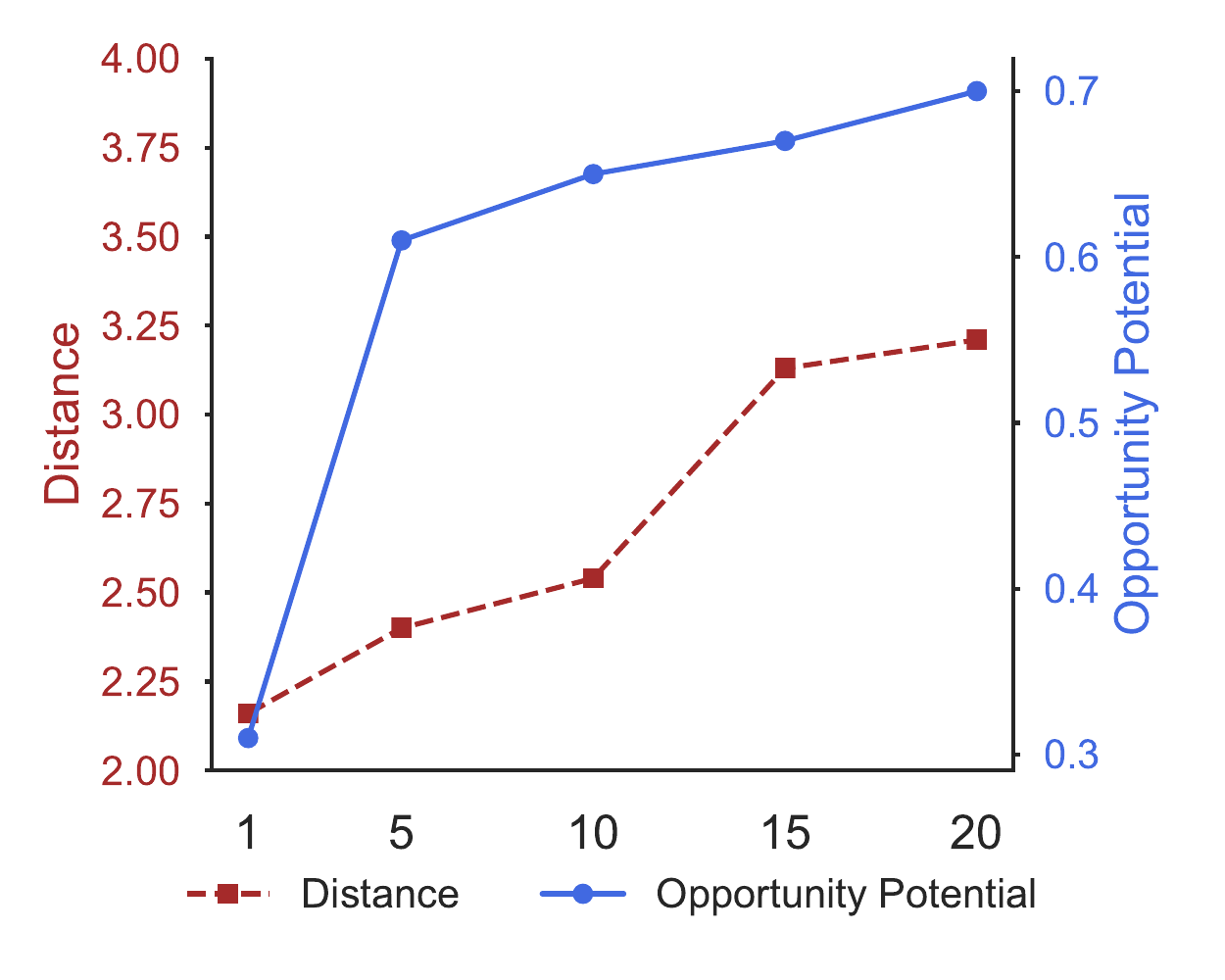}
        \caption{Impact of $c$ in German Credit.}
    \end{subfigure}
    \vspace{0.5em}%
    \begin{subfigure}[b]{0.325\textwidth}%
        \includegraphics[width=\textwidth]{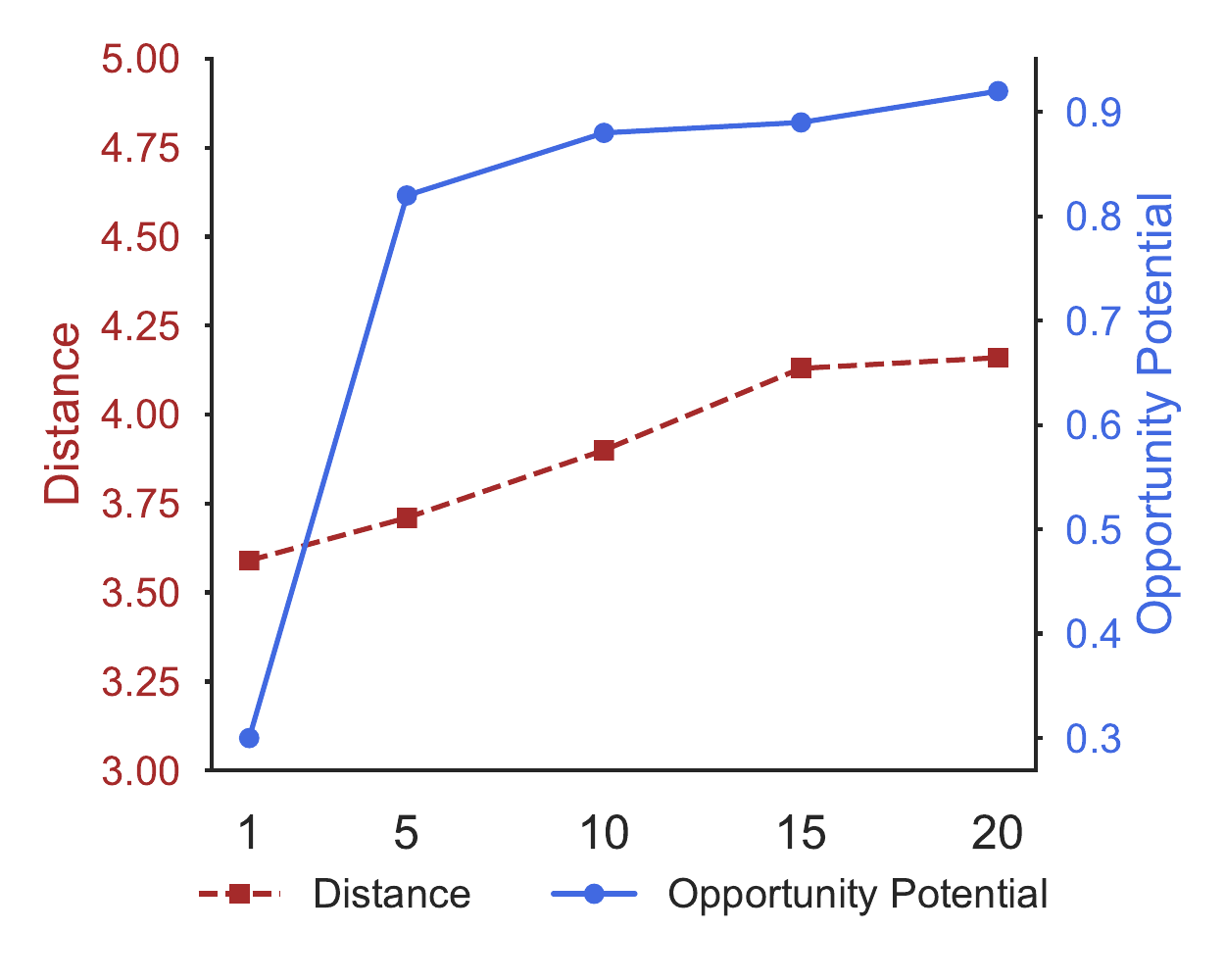}
        \caption{Impact of $c$ in Adult Income.}
    \end{subfigure}
    \caption{Change of \emph{distance} and \emph{opportunity potential} of the most optimal \textsc{FACElift} counterfactual as $c$ increases.}%
    \label{fig:ablation}
\end{figure}

To better understand \textsc{FACElift}'s behaviour we scrutinise its parametrisation; %
specifically, we focus on %
the number of (closest) counterfactual candidates $c$ from which our explainer selects the one with the highest opportunity potential. %
By design, increasing $c$ yields a larger pool of counterfactuals, thus improving the chances of identifying one with superior opportunity potential. %
However, doing so also makes \textsc{FACElift} more likely to output an explanation that is farther away from the explained (factual) instance. %
To assess this trade-off, we vary $c$ for the German Credit and Adult Income data sets while comparing %
\emph{opportunity potential} and \emph{distance} of optimal counterfactuals returned by \textsc{FACElift}. %
The results, shown in Figure~\ref{fig:ablation}, %
demonstrate that increasing $c$ consistently improves opportunity potential -- the benefit of having access to %
a broader pool of candidate explanations. %
This, nonetheless, %
comes at the expense of larger distance as the extended set of counterfactuals includes instances that are farther away from the factual point. %
However, %
with $c$ increasing, %
the boost in opportunity potential quickly diminishes as distance %
continues to grow. %
Based on our findings, we recommend keeping $c$ relatively small to effectively manage the trade-off between maximising user agency -- as measured by opportunity potential -- and minimising recourse cost -- as measured by distance. %

\section{Discussion and Future Work\label{sec:discussion}}

Explanatory multiverse is a novel conceptualisation of step-based counterfactual explanations (overviewed in Sections~\ref{sec:intro} and \ref{sec:prelim:related}) that inherits all of their desired, spatially-unaware properties; %
it further %
extends them with a collection of spatially-aware desiderata that take advantage of the geometry of counterfactual paths (covered in Section~\ref{sec:prelim:desiderata}). %
Despite being overlooked in the technical literature, such a view on counterfactual reasoning, explainability and decision-making is largely consistent with the account of these domains in philosophy, psychology and cognitive sciences~\cite{kahneman2014varieties,mueller2021principles}. %

\citeauthor{lewis1973counterfactuals}'~(\citeyear{lewis1973counterfactuals}) notion of \emph{possible-worlds}, of which the real world is just one and with some being closer to reality (i.e., more realistic) than others~\cite{kahneman1990propensities}, is akin to the numerous interrelated paths connecting the factual instance with counterfactual explanations captured by our formalisation of explanatory multiverse. %
Specifically, it accounts for spatial and temporal contiguity, thus supports and better aligns with \emph{mental simulation}, which is a form of automatic and elaborative (counterfactual) thinking. %
This process spans continuum rather than a typology and allows humans to imagine congruent links between two possible states of the world, e.g., factual and counterfactual, by unfolding a sequence of events connecting them. %
Additionally, explanatory multiverse supports complex human reasoning and decision-making more effectively than (explainable) ML systems that simply output (and justify) a single prediction that is the most optimal according to the underlying optimisation objective. %
It encourages and helps %
people to consider multiple possible courses of action -- specifying their respective key branching points or critical junctures -- %
thus allowing explainees to %
develop effective strategies and action plans that remain valid %
for various potential outcomes and can be %
easily adapted to unexpected developments~\citep{helmer1959epistemology}. %

Navigating the nexus of (hypothetical) possibilities through the (technical) proxy of explanatory multiverse enables structured -- and possibly (interactively) guided -- exploration, comparison and reasoning over (remote branches of) counterfactual explanations. %
By accounting for the spatial and temporal aspects of their paths -- such as affinity, branching, divergence and convergence -- we construct an explainability framework capable of modelling a perceptual distance between counterfactual trajectories. %
For example, it allows us to consider people's tendency to first undo the changes implemented at the beginning of a process and their propensity to fall prey to the sunk cost fallacy, i.e., continuing an endeavour after the initial effort~\cite{arkes1985psychology}. %
To the best of our knowledge, we are the first to model the geometry of explanations, thus better align ML interpretability with modes of counterfactual thinking found in humans. %

Explanatory multiverse also %
advances the human-centred explainability agenda on multiple fronts. %
It comes with an inherent heuristic to deal with counterfactual multiplicity by recognising their spatial (dis)similarity, thus reducing the (possibly overwhelming~\citep{herm2023impact}) cognitive load required of explainees. %
By lowering the choice complexity when deciding on actions to take while navigating and traversing step-based counterfactual paths, explanatory multiverse better aligns this task with iterative and interactive, dialogue-based, conversational explainability -- a process that is second nature to humans~\cite{miller2019explanation,keenan2023mind}. %
The increased agency, however, %
may in itself inadvertently contribute to cognitive overload. %
Appropriate parameterisation of our approach can help to mitigate such undesired side effects; %
by controlling how spatially (dis)similar explanations should be grouped together (refer to the \emph{direction difference} metric discussed in Section~\ref{sec:vector}), %
we can manage the inherent similarity--complexity trade-off. %
This setting can also be modified dynamically %
to, at first, generate a high-level overview of available explanations, %
before drilling down into the details as users progress along counterfactual journeys (refer to the \emph{path branching} discussion in Section~\ref{sec:graph} for an illustrative example). %

Given the ability to consider the general (geometrical) direction of representative counterfactual paths instead of their specific instantiations -- which may be overwhelming due to their quantity and lack of meaningful differentiation -- our conceptualisation is compatible with both the well-established \emph{justification} as well as the more recent \emph{decision-support} paradigms of explainable ML~\cite{miller2023explainable}. %
An additional benefit of explanatory multiverse is its ability to uncover disparity in access to counterfactual recourse -- a fairness perspective; %
some individuals may only be offered a limited set of explanations, or none at all, if they belong to remote clusters of points, e.g., portraying a protected group that is underrepresented in the data. %

The three preliminary desiderata outlined in Section~\ref{sec:prelim:desiderata} are general enough to encompass most applications of explanatory multiverse, nonetheless they are far from exhaustive; %
we envisage identifying more properties as we explore this concept further and apply it to specific problems. %
Doing so will allow us to learn about their various trade-offs, especially since we expect different data domains and modelling problems to prioritise distinct desiderata. %
Forgoing na{\"i}ve optimisation for the shortest path in favour of a deliberate detour can benefit explainees on multiple levels as argued throughout this paper. %
Some of these considerations can be communicated through intuitive visualisations, making them more accessible (to a lay audience); %
e.g., we may plot the number of implemented changes against the proportion of counterfactual recourse opportunities that remain available at any given point. %

With the current set of properties, for example, one may prefer delayed branching, which incurs small loss of opportunity early on but a large one at later stages, thus initially preserving high agency. %
Similarly, a small loss of opportunity early on can be accepted to facilitate delayed branching, hence reduce overall choice complexity at the expense of agency. %
If, in contrast, early branching is desired -- e.g., one prefers a medical diagnosis route with fewer required tests -- this will result in a large loss of opportunity at the beginning, thus lower agency. %
Additionally, certain paths may be easier to follow, i.e., preferred, due to domain-specific properties, e.g., a non-invasive medical examination, %
whereas others may require crossing points of no return, i.e., implementing changes that cannot be (easily) undone. %

Furthermore, %
explanatory multiverse is %
intrinsically compatible with \emph{multi-class counterfactuals} -- refer back to the MNIST example shown in Figure~\ref{fig:mnist} -- as well as \emph{probabilistic explainability}~\cite{sokol2020limetree,sokol2025all}; %
these paradigms offer a nuanced and comprehensive explanatory perspective that is unavailable for other explanation types such as feature importance, feature influence or exemplars. %
For instance, %
note that an action may have multiple distinct consequences: make certain outcomes more likely (or simply possible), other less likely (or even impossible), or both at the same time. %
Therefore, depending on the point of view, each step taken by an explainee can be interpreted as a negative or positive (counterfactual) explanation. %
Our approach also facilitates \emph{retrospective explanations} that allow to (mentally) \emph{backtrack} steps leading to the current situation -- in contrast to \emph{prospective explanations} that prescribe actionable insights -- allowing to answer questions such as ``How did I end up here?'' %

In view of the sheer potential of our approach and the promising results offered by our experiments, streamlining and extending %
diverse aspects of explanatory multiverse are important next steps. %
Specifically, %
the vector space conceptualisation %
could benefit from more comprehensive and versatile %
definitions of selected measures and properties as well as efficient and universal implementations (e.g., based on the uncertainty estimates or gradients output by predictive models). %
As mentioned earlier, this family of methods %
appears best suited for continuous feature spaces for which a sizeable and representative sample of data is available. %
The graph realisation and implementation, on the other hand, %
could be better aligned with the vector space formalisation and %
gain from various algorithmic optimisations; %
overall, methods of this type %
seem highly appropriate for somewhat sparse data sets with mostly discrete attributes as noted earlier. %
Other realisations and implementations of explanatory multiverse as well as additional spatially-aware properties and metrics are also worth exploring. %
All of these developments %
will allow us to transition away from relatively simple scenarios and apply our tools to more complex, real-life data sets. %

In addition to advancing our methodology introduced in this paper, %
in future work we will study building \emph{dynamical systems}, \emph{phase spaces} and \emph{vector fields} from (partial) \emph{sequences} of observations to capture complex dynamics, e.g., divergence, turbulence, stability and vorticity, within explanatory multiverse. %
Furthermore, we will extend our techniques to %
structured data such as \emph{ordered events} and \emph{time series} %
-- building upon appropriate pre-existing explainers to this end~\citep{beretta2023importance,de2024time} -- %
since explanatory multiverse %
fits naturally in these scenarios. %
Both of these developments appear particularly promising for %
medical (electronic health record) data. %
We will also look into \emph{explanation representativeness} -- i.e., identifying and grouping counterfactuals that are the most diverse and least alike -- given that it is central to navigating the geometry of counterfactual paths yet largely under-explored. %
In particular, discovering (dis)similarity of counterfactuals can streamline the exploration of explanatory multiverse, %
helping to identify pockets of highly attractive or inaccessible explanations. %
Understanding these dependencies is also important given their \emph{fairness ramifications} as some individuals may only have limited (counterfactual) recourse options. %
Finally, we plan to investigate explanatory multiverse from a \emph{causal viewpoint} %
as well as scrutinise the impact of \emph{model correctness} on the quality of our explanations since these two perspectives are often neglected~\citep{sokol2024what}. %

\section{Conclusion\label{sec:conclusion}}

In this paper we introduced \emph{explanatory multiverse}: a novel conceptualisation of counterfactual explainability that takes advantage of \emph{geometrical relation} -- affinity, branching, divergence and convergence -- between paths representing the steps connecting factual and counterfactual data points. %
Our approach better aligns such explanations with human needs and expectations by distilling informative, intuitive and actionable insights from, otherwise overwhelming, \emph{counterfactual multiplicity}; %
explanatory multiverse is also compatible with iterative and interactive explanatory protocols, which are a key tenet of human-centred explainability. %

To guide the retrieval of high-quality explanations, we formalised three \emph{spatially-aware desiderata}: agency, loss of opportunity and choice complexity; %
nonetheless, this is just a preliminary and non-exhaustive set of properties that we expect to expand as we further explore explanatory multiverse and apply it to distinct data domains. %
In addition to foundational and theoretical contributions %
conceptualised with vector spaces %
-- which we showcased on a synthetic two-dimensional tabular data set -- %
we proposed an explanatory multiverse implementation %
based on (directed) graphs. %
We used the latter to execute experiments on six diverse data sets spanning tabular and image domains, %
demonstrating real-life capabilities of our technique. %
We also introduced a high-level flexible evaluation metric -- called \emph{opportunity potential} -- that encompasses all of our spatially-aware desiderata; %
we then used it %
to quantitatively verify the benefit and efficacy of our approach. %
Notably, %
our method, the open source implementation of which is available on GitHub to promote reproducibility, is built upon a state-of-the-art step-based counterfactual explainer %
-- called FACE -- %
therefore it comes equipped with all the desired spatially-unaware properties out of the box. %

\renewcommand{\acksname}{Acknowledgements}
\begin{acks}
This research was conducted by the ARC Centre of Excellence for Automated Decision-Making and Society (project number CE200100005), funded by the Australian Government through the Australian Research Council. %
Additional support was provided by the Hasler Foundation (grant number 23082). %
\end{acks}

\section*{Conflicts of Interest}%
We declare no competing interests.

\section*{Data Transparency}
\begin{description}[labelindent=0cm,labelwidth=1em,labelsep=1em,align=left,font=\it]%
\setlength\itemsep{0em}
\item [German Credit:]
\url{https://archive.ics.uci.edu/dataset/144/statlog+german+credit+data}
\item [Adult Income:]
\url{https://archive.ics.uci.edu/dataset/2/adult}
\item [Credit Default:]
\url{https://archive.ics.uci.edu/dataset/350/default+of+credit+card+clients}
\item [MNIST:]
\url{http://yann.lecun.com/exdb/mnist}
\item [MedMNIST:]
\url{https://medmnist.com}
\end{description}

\section*{Code Availability}
\sourcecode

\section*{Author Contribution}
\begin{description}[labelindent=0cm,labelwidth=1em,labelsep=1em,align=left,font=\it]%
\setlength\itemsep{0em}
\item [Conceptualisation:]
KS (lead investigator), ES, YX %
\item [Methodology:]
KS (explanatory multiverse), ES (vector-based approach), YX (graph-based approach) %
\item [Code development:]
ES (vector-based approach), YX (graph-based approach) %
\item [Writing:]
KS (explanatory multiverse), ES (vector-based approach), YX (graph-based approach) %
\item [Review and editing:]
KS
\end{description}

\bibliographystyle{ACM-Reference-Format}
\bibliography{arxiv}

\appendix

\section{Closest Point to Parameterised Line Proof\label{apx:cose_point}}

Given three points $x_1$, $x_2$ and $x_3$, we seek the closest point to $x_3$ on the straight line defined by $x_1$ and $x_2$. %
The parameterised line that connects $x_1$ and $x_2$ is defined as %
\begin{equation*}
    f(l) = (1-l)x_1 + lx_2
    \text{~.}
\end{equation*}
For $l=0$, $f(l)=x_1$; and for $l=1$, $f(l)=x_2$.
Therefore, we have an equation where the values of $l\in[0,1]$ represent points between $x_1$ and $x_2$.

Next, we define a vector between the parameterised line and the point of interest $x_3$ as %
\begin{equation*}
    g(l) = f(l) - x_3 = (1-l)x_1 + lx_2 - x_3
    \text{~.}
\end{equation*}
Our goal is to find the value of $l$ that minimises the length of this vector such that the distance between $x_3$ and $f(l)$ is the shortest.
The squared length of this vector is
\begin{equation*}
    \lVert g(l) \rVert ^2 = (l(x_2 - x_1) + x_1 - x_3)^T \cdot (l(x_2 - x_1) + (x_1 - x_3))
    \text{~.}
\end{equation*}
We represent the difference between the points of interest as vectors
\begin{equation*}
    w = x_2 - x_1 \qquad \text{and} \qquad u = x_1 - x_3
    \text{~,}
\end{equation*}
which gives
\begin{equation*}
\begin{aligned}
    \lVert g(l) \rVert ^2 &= (lw + u)^T \cdot (lw + u) \\
                          &= l^2\lVert{w}\rVert^2 + 2l(w\cdot u) + \lVert u \rVert ^2
    \text{~.}
    \end{aligned}
\end{equation*}
Finding $\nabla \lVert g(l) \rVert ^2 = 0$ gives the solution:
\begin{equation*}
    l = - \frac{w \cdot u}{w \cdot w}
    \text{~.}
\end{equation*}

\section{Opportunity Potential Examples\label{apx:examples}}%

\begin{figure*}[t]%
    \centering
    \begin{subfigure}[b]{0.485\textwidth}
         \centering
         \includegraphics[height=2.85cm]{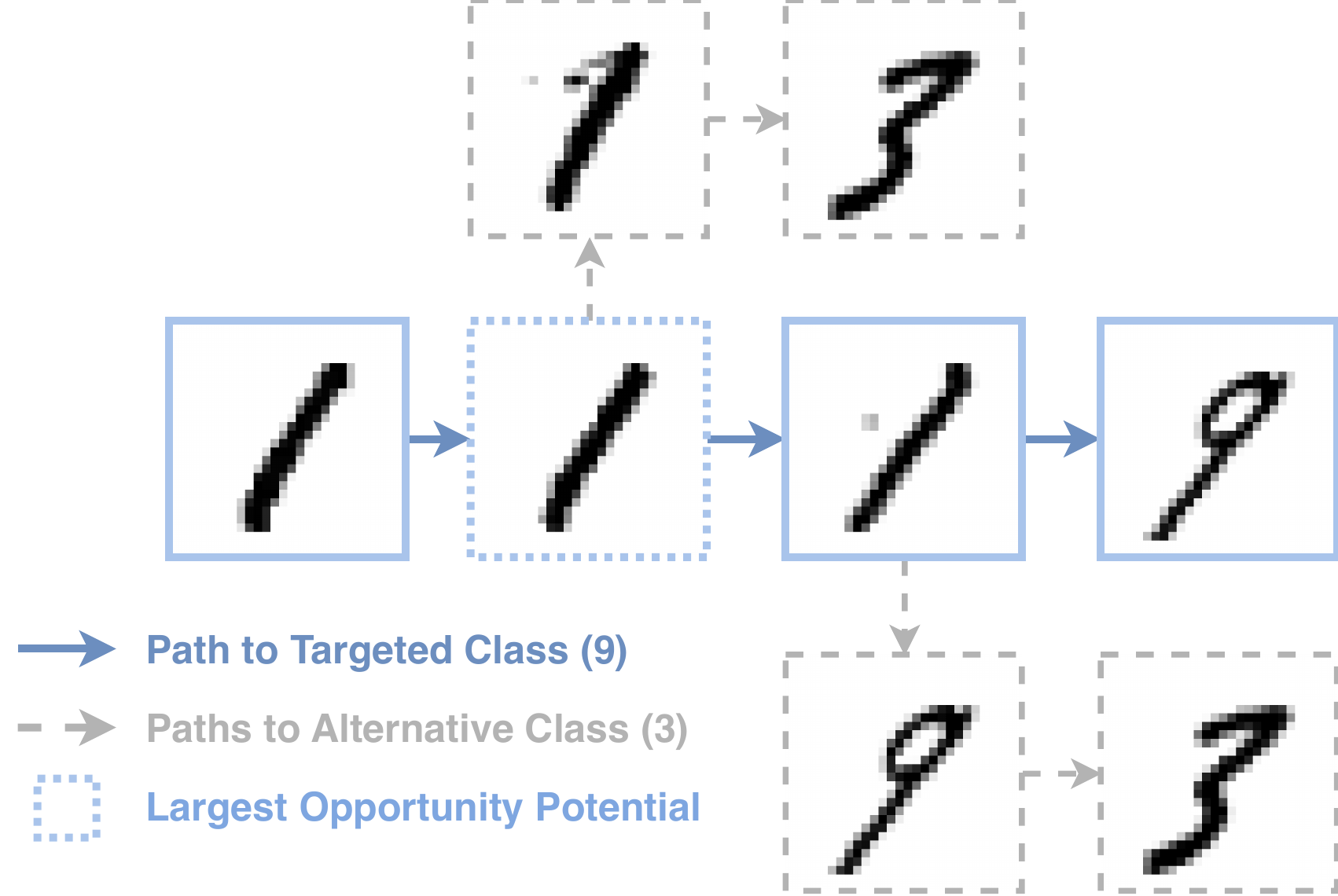}%
         \caption{The shortest path has overall \emph{opportunity potential} of 0.14.}%
         \label{fig:mnist-cal-a}
    \end{subfigure}
    \hfill
    \begin{subfigure}[b]{0.485\textwidth}
         \centering
         \includegraphics[height=2.85cm]{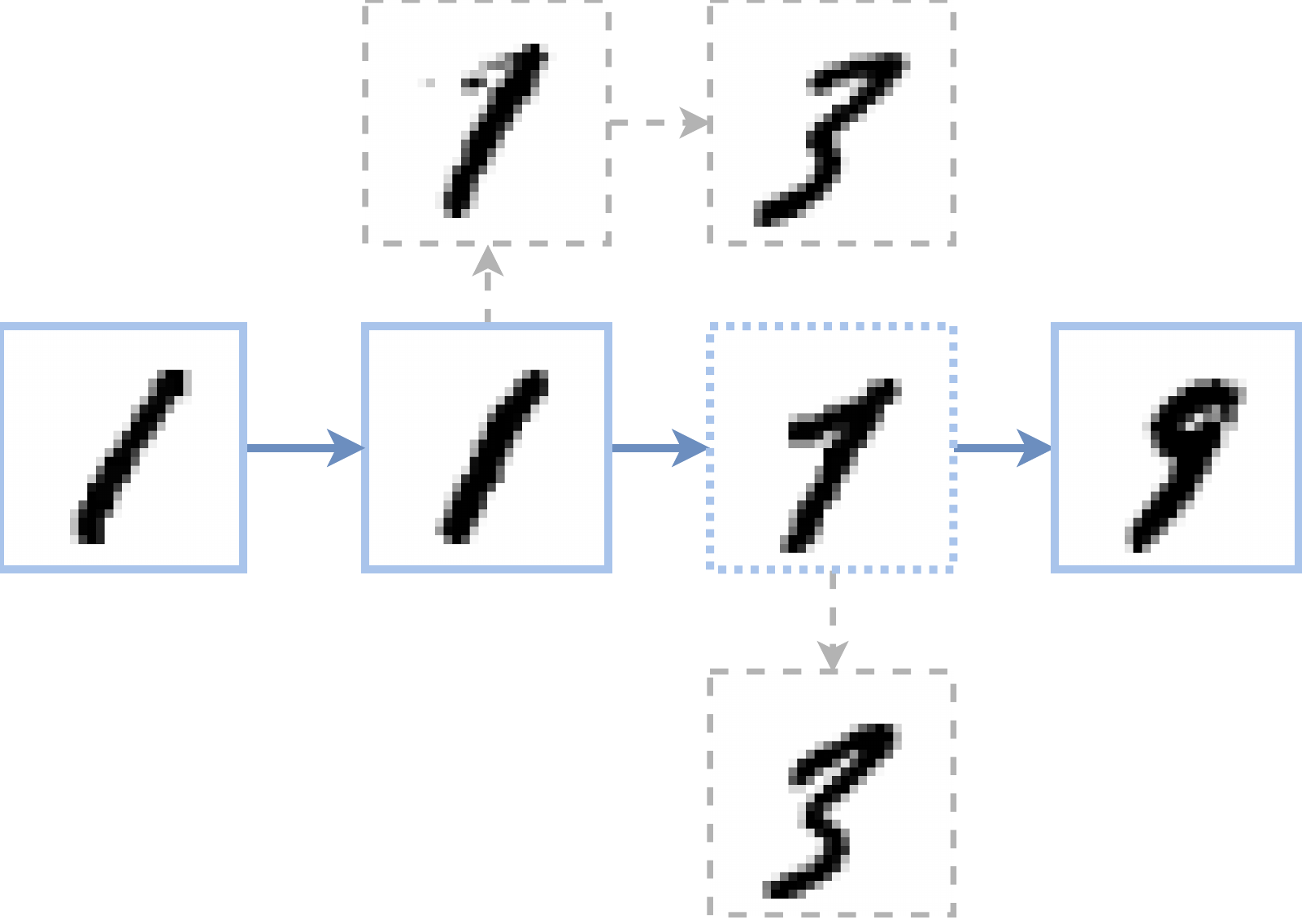}%
         \caption{The 2\textsuperscript{nd} shortest path has overall \emph{opportunity potential} of 0.58.}%
         \label{fig:mnist-cal-b}
     \end{subfigure}
     \\[1em]
    \begin{subfigure}[b]{0.485\textwidth}
         \centering
         \includegraphics[height=2.85cm]{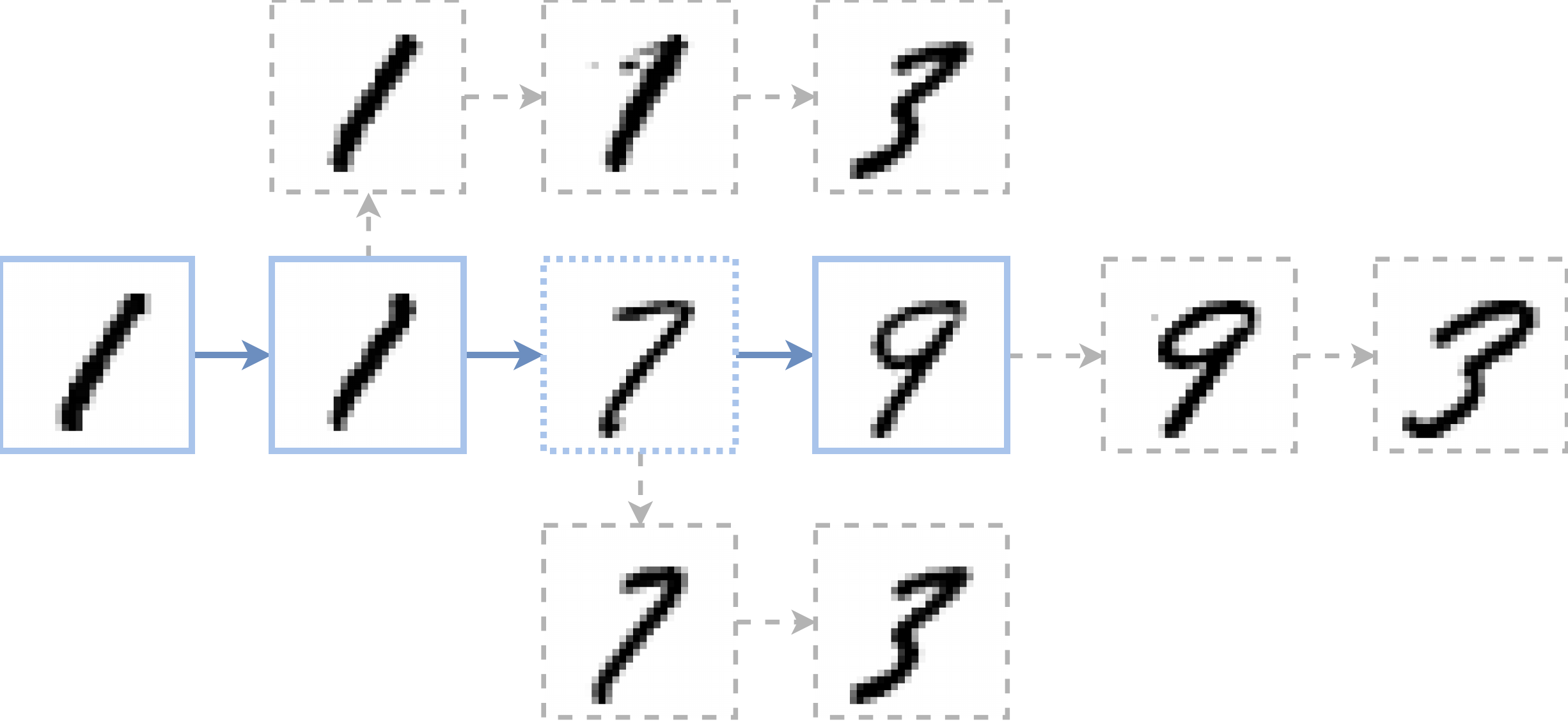}%
         \caption{The 3\textsuperscript{rd} shortest path has overall \emph{opportunity potential} of 0.62.}%
         \label{fig:mnist-cal-c}
    \end{subfigure}
    \hfill
    \begin{subfigure}[b]{0.485\textwidth}
         \centering
         \includegraphics[height=2.85cm]{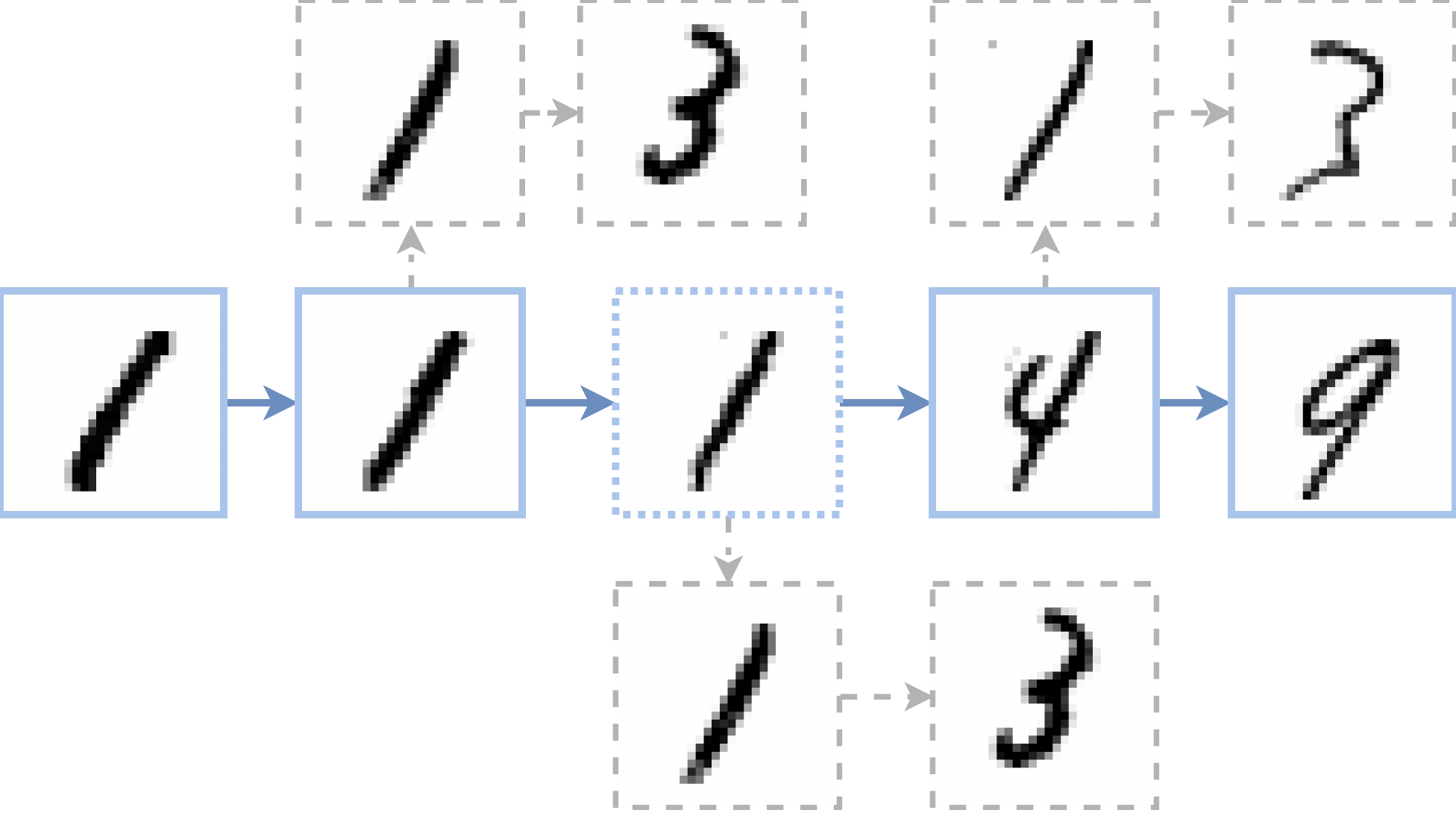}%
         \caption{The 4\textsuperscript{th} shortest path has overall \emph{opportunity potential} of 0.35.}%
         \label{fig:mnist-cal-d}
    \end{subfigure}
    \caption{%
    Four different counterfactual paths targeting class 9 for the same factual instance of class 1. %
    Due to space limitation, we only show alternative paths that lead to counterfactuals of class 3. %
    }%
    \label{fig:mnist_calculation}%
\end{figure*}

Figure~\ref{fig:mnist_calculation} shows the top four (optimal) counterfactual paths ($\check{y} = 9$) as well as paths to an alternative class ($\check{y} = 3$) for a single instance ($\mathring{y} = 1$) taken from the MNIST data set. %
Figure~\ref{fig:mnist-cal-a} depicts the shortest path, but its opportunity potential is low; past the second step the path begins to diverge from $\check{y} = 3$, i.e., digit 3 classification. %
The paths displayed in Figures~\ref{fig:mnist-cal-b} and \ref{fig:mnist-cal-c} are longer but have higher opportunity potential. %
Figure~\ref{fig:mnist-cal-d} depicts the longest path whose opportunity potential is lower than that of the previous two paths, making it sub-optimal. %
These examples further demonstrate the inherent trade-off between the length and opportunity potential of a counterfactual path. %

\section{Post-hoc Counterfactual Path Construction via Binary Space Partition}\label{app:bsp}%

While recent research introduced numerous sequential counterfactual explainers, none of these approaches except for FACE is compatible with image data~\cite{poyiadzi2020face,ramakrishnan2020synthesizing,kanamori2021ordered,verma2022amortized}. %
To adapt generic counterfactual explainers to explanatory multiverse we propose a Binary Space Partition (BSP) approach -- formalised in Algorithm~\ref{alg:bsp} -- that constructs counterfactual paths for a given explanation post-hoc. %
Such a counterfactual path is composed of a sequence of existing data instances that are located between the factual and counterfactual points.
Specifically, for a given counterfactual instance we recursively partition the space between the factual and counterfactual instances at midpoints until no intermediate data points are left within these partitions or their size is smaller than a predefined threshold. %

\begin{algorithm}%
\caption{%
Construct paths for a counterfactual explanation using Binary Space Partition. %
}%
\label{alg:bsp}%
\begin{algorithmic}[1]
    \REQUIRE
    factual point $\mathring{x}$;
    counterfactual point $\check{x}$;
    data set $X$;
    partition size threshold $\tau$. %
    \ENSURE
    counterfactual path $Z$.
\STATE \textbf{function} \texttt{BSP}($X$, x\_indices, $x_a$, $x_b$, $d$)
\begin{ALC@g}
    \STATE dist\_a $\gets ||X, x_a||_2$
    \COMMENT{compute $L^2$ distance between $x_a$ and every point in $X$}%
    \STATE dist\_b $\gets ||X, x_b||_2$
    \STATE sub\_indices $\gets$ \textit{arg}(dist\_a $\leq d$ \&\& dist\_b $\leq d$)
    \RETURN x\_indices[sub\_indices]
\end{ALC@g} 
\STATE \textbf{function} \texttt{recursiveBSP}($X$, x\_indices, $x_i$, $x_j$, $d$, $\tau$)
\begin{ALC@g}
    \IF{$d<\tau$}
    \STATE x\_id $\gets$ \textit{random\_choice}(x\_indices) \COMMENT{randomly select an instance}
    \RETURN x\_id
    \ENDIF
    \STATE mid\_point $\gets (x_i+x_j)/2$
    \COMMENT{find the point splitting the space in half}
    \STATE $d \gets d-1$
    \STATE partition\_l $\gets$ \texttt{BSP}($X$, x\_indices, $x_i$, mid\_point, $d$)
    \STATE partition\_r $\gets$ \texttt{BSP}($X$, x\_indices, mid\_point, $x_j$, $d$)
    \IF{partition\_l == $\emptyset$ \;\&\&\; partition\_r == $\emptyset$}
    \STATE x\_id $\gets$ \textit{random\_choice}(x\_indices)
    \RETURN x\_id
    \ELSE
        \IF{partition\_l != $\emptyset$}
            \STATE x\_indices\_l $\gets$ \texttt{recursiveBSP}($X$, partition\_l, $x_i$, mid\_point, $d$, $\tau$)
        \ELSE
            \STATE x\_indices\_l $\gets \emptyset$
        \ENDIF
        \IF{partition\_r != $\emptyset$}
            \STATE x\_indices\_r $\gets$ \texttt{recursiveBSP}($X$, partition\_r, mid\_point, $x_j$, $d$, $\tau$)
        \ELSE
            \STATE x\_indices\_r $\gets \emptyset$
        \ENDIF
    \ENDIF
    \RETURN x\_indices\_l + x\_indices\_r
\end{ALC@g}
\STATE \textbf{function} \texttt{findPath}($X$, $\mathring{x}$, $\check{x}$, all\_x\_indices, $\tau$)%
\begin{ALC@g}
    \STATE init\_d $\gets ||\check{x}, \mathring{x}||_2$ 
    \STATE path\_points\_indices $\gets$ \texttt{recursiveBSP}($X$, all\_x\_indices, $\mathring{x}, \check{x}$, init\_d, $\tau$)
    \STATE path\_points\_indices $\gets$ \textit{sort}(path\_points\_indices)
    \COMMENT{sort indices of points based on their distance to $\check{x}$}
    \RETURN path\_points\_indices
\end{ALC@g}
\STATE all\_x\_indices $\gets [0, \ldots, |X|]$
\STATE
$Z \gets$ \texttt{findPath}($X$, $\mathring{x}$, $\check{x}$, all\_x\_indices, $\tau$)
\end{algorithmic}
\end{algorithm}

The counterfactual path generated by BSP is guaranteed to monotonically approach the target instance at every step. %
Setting the size threshold prevents partitions from becoming too small, which ensures that the counterfactual path does not consist of steps that are too close to each other to remain meaningful. %
On MNIST, the average number of steps in a counterfactual path generated by BSP is $5.21 \pm 1.45$ for GrowingSphere and
$5.64 \pm 1.80$ for Prototype. %
For comparison, FACE is a path-based explainer that directly outputs counterfactual paths whose number of steps is $3.37 \pm 0.69$. %
Therefore, BSP performs on a par with a state-of-the-art path-based explainer. %

\end{document}